\documentclass[a4paper,fleqn]{cas-dc}
\usepackage{float}
\usepackage[numbers]{natbib}
\usepackage{xcolor} 
\usepackage{tabularx}
\newcommand{\revise}[1]{{\textcolor{black}{#1}}}

\def\tsc#1{\csdef{#1}{\textsc{\lowercase{#1}}\xspace}}
\tsc{WGM}
\tsc{QE}
\tsc{EP}
\tsc{PMS}
\tsc{BEC}
\tsc{DE}

\usepackage{xcolor}

\begin{document}
\let\WriteBookmarks\relax
\def\floatpagepagefraction{1}
\def\textpagefraction{.001}

\shorttitle{Chart Specification}

\shortauthors{Minggui He et al.}

\title [mode = title]{Chart Specification: Structural Representations for Incentivizing VLM Reasoning in Chart-to-Code Generation}

\author[1]{Minggui He}
\author[2]{Mingchen Dai}

\author[1]{Jian Zhang}
\author[3]{Yilun Liu}
\author[2]{Shimin Tao}
\author[2]{Pufan Zeng}
\author[1]{Osamu Yoshie} 
\cortext[cor1]{Corresponding author}
\ead{yoshie@waseda.jp}
\author[1]{Yuya IEIRI}

\fntext[eq]{Minggui He and Mingchen Dai contributed equally to this work.}

\affiliation[1]{organization={Waseda University},
    country={Japan}}
\affiliation[2]{organization={University of Science and Technology of China},
    country={China}}
\affiliation[3]{organization={NanKai University}, country={China}}

\begin{abstract}
Vision-Language Models (VLMs) have shown promise in generating plotting code from chart images, yet achieving structural fidelity remains challenging. Existing approaches largely rely on supervised fine-tuning, encouraging surface-level token imitation rather than faithful modeling of chart structure, which often leads to hallucinated or semantically inconsistent outputs.
We propose Chart Specification, a canonical structural representation that shifts training from mimicking training-code patterns to structure-grounded learning. By normalizing plotting code into structure-equivalent specifications, it enables (i) the construction of a structurally balanced training set, and (ii) a Spec-Align Reward that provides fine-grained, verifiable feedback on structural correctness for reinforcement learning. Under an explicit reasoning-to-code generation paradigm, this reward encourages structure-aware reasoning that produces constraint-consistent plotting code.
Experiments on three public benchmarks show that our method consistently outperforms prior approaches. With only 3K training samples, we achieve strong data efficiency, surpassing leading baselines by up to 61.7\% on complex benchmarks, and scaling to 4K samples establishes new state-of-the-art results across all evaluated metrics. Overall, our results demonstrate that precise structural supervision offers an efficient pathway to high-fidelity chart-to-code generation. Code and dataset are available at: \url{https://github.com/Mighten/chart-specification-paper}

\end{abstract}


\begin{highlights}
\item \revise{We introduce Chart Specification, a canonical structural representation for faithful chart-to-code generation.}

\item \revise{It normalizes plotting code into structure-equivalent specifications to curate a structurally balanced training set.}

\item \revise{We design a Spec-Align Reward that provides fine-grained, verifiable structure-level feedback to incentivize VLM reason-then-code generation via reinforcement learning.}

\item \revise{ Extensive experiments on three public benchmarks demonstrate state-of-the-art performance with strong data efficiency.}

\end{highlights}

\maketitle

\section{Introduction}
Charts constitute a primary medium for quantitative communication, conveying dense numerical values, structural relationships, and analytical intent across the scientific literature, business intelligence dashboards, and public reports.
As a result, the ability to automatically interpret and repurpose chart visualizations is a fundamental requirement for intelligent document processing systems~\cite{siegel2016figureseer}.
Among related tasks, \emph{chart-to-code generation} has recently emerged as a particularly challenging and consequential problem.
Given a static raster chart image, the goal is to generate executable plotting code that faithfully reconstructs the original figure~\cite{poco2017reverse}.
Beyond practical applications such as chart editing, reuse, and reproducibility, this task serves as a strict test of a model’s ability to recover layout structure, data mappings, and numerical relationships from visually encoded information~\cite{kafle2018dvqa}.

Recent advances in Large Vision-Language Models (VLMs) suggest a promising direction for addressing this challenge.
Pre-trained on large-scale image–text corpora, VLMs demonstrate strong capabilities in visual grounding and cross-modal alignment, making them natural candidates for mapping chart images to high-level plotting instructions~\cite{zhang2024vision}. 
However, despite these advances, robust chart-to-code generation remains elusive.
Unlike natural images, charts encode precise and structured information, such as bar heights, line trajectories, and spatial dependencies, where even minor errors in code can lead to significant visual discrepancies.
Although recent benchmarks and datasets~\cite{yang2024chartmimic,wu2024plot2code} have begun to explore this problem, existing approaches still exhibit brittle generalization and persistent structural errors.

\begin{figure*}[tb]
    \centering

    \includegraphics[width=\textwidth]{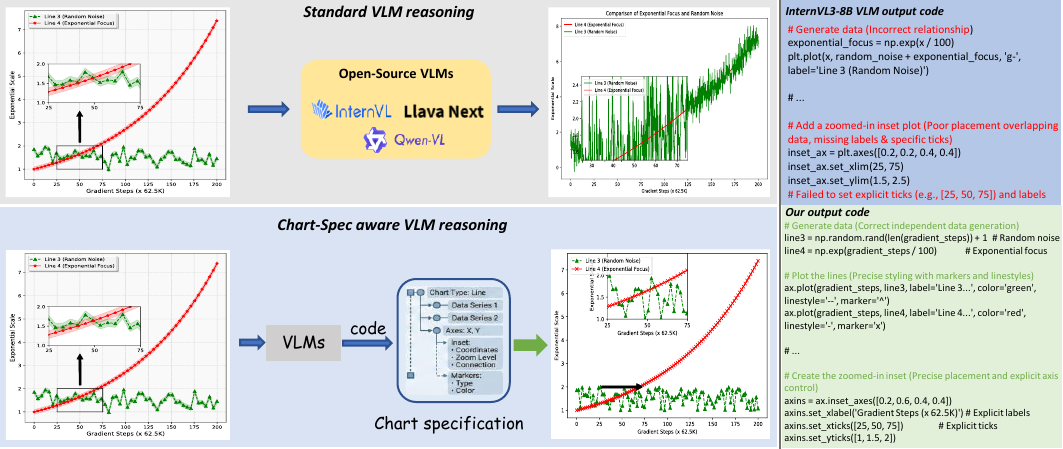}
    
    \caption{\textbf{Motivation for structure-aware chart reasoning.}
    (Top) Direct chart-to-code models rely on surface-level imitation and often hallucinate structural dependencies.
    (Bottom) By explicitly modeling chart structure via Chart Specification, our approach enforces constraint-consistent plotting logic and faithful visual reconstruction.}
    \label{fig:motivation}
\end{figure*}

We argue that these failures stem from a fundamental limitation of \emph{direct chart-to-code supervision}.
Charts encode information in a continuous and spatially structured visual domain, while the plot code is discrete, symbolic, and procedural.
Directly supervising code tokens from chart images forces models to bridge this cross-domain gap in a single step, where the supervision signal is dominated by syntactic variability rather than visual or structural intent.
As a result, token-level objectives provide weak guidance on which visual structures must be preserved, leading to unstable optimization and poor structural fidelity.
This mismatch gives rise to three characteristic failure modes observed in existing chart-to-code systems:

\textbf{First, models are prone to structural hallucinations due to the dual burden of visual reasoning and code syntax.}
Unlike conventional VLM tasks, chart-to-code generation requires precise interpretation of visual elements while simultaneously composing valid plotting commands.
As illustrated in Figure~\ref{fig:motivation}, a baseline model hallucinates a dependency between independent data series, incorrectly encoding "Random Noise" as an additive component of the "Exponential Focus" trend.
Although the generated code is syntactically valid, it fails to reflect the true visual semantics of the chart.

\textbf{Second, the token-level prediction objective is inherently misaligned with the goal of visual fidelity.}
Existing approaches, such as ChartCoder~\cite{zhao2025chartcoder}, primarily rely on scaling supervised training data to improve performance.
However, next-token prediction treats plotting code as a sequence of text tokens, largely ignoring the visual consequences of individual commands.
As a result, severe numerical or layout errors are weakly penalized as long as the generated code remains statistically plausible, preventing models from developing global structural awareness.

\textbf{Third, the lack of outcome-oriented feedback limits further optimization.}
While recent work has explored reinforcement learning (RL) to improve executable code quality~\cite{lightman2023letsverify,le2022coderl}, applying RL to chart-to-code generation exposes a critical reward gap.
Binary execution feedback is overly sparse, as many structurally incorrect charts compile successfully, whereas pixel-level comparisons are unreliable due to noise from stylistic and non-structural variations.
Without a verifiable and fine-grained metric that reflects structural correctness, it is infeasible to guide models toward consistent and constraint-respecting plotting logic.

To overcome these limitations, we propose \textbf{Chart Specification}, a canonical structural representation that explicitly encodes the structural logic of a chart.
Rather than treating plotting code as plain text for token-level imitation, Chart Specification abstracts essential visual factors, such as layout composition, coordinate systems, data bindings, and functional relationships, while remaining invariant to superficial syntactic differences in code.
Crucially, this representation is aligned with both the visual domain and the execution semantics of plotting libraries, enabling supervision signals to be transmitted more effectively across modalities.

Building on the Chart Specification, we design two complementary components that operationalize this representation during training.
First, we construct \textbf{ChartStruct}, a structurally balanced dataset that addresses the long-tail distribution hidden within coarse chart categories.
By organizing samples according to fine-grained specifications, ChartStruct ensures coverage of diverse generative patterns, such as line charts defined by explicit functional mappings versus those derived from discrete interpolated data.
Second, we introduce the \textbf{Spec-Align Reward}, a structure-aware metric that directly measures the agreement between generated and reference specifications.
This reward provides dense and verifiable feedback suitable for reinforcement learning, effectively closing the reward gap in chart-to-code optimization.
We adopt an explicit reason-then-code generation paradigm to encourage the VLM to first produce a reasoning trace and then synthesize plotting code. we compute rewards by converting the generated code into Chart Specification and comparing it against the reference specification.

We evaluate our approach on three public chart-to-code benchmarks.
Despite using substantially fewer training samples, our method consistently outperforms prior approaches on complex benchmarks with only 3K samples, and further establishes new state-of-the-art results when scaled to 4K.
These results demonstrate that Chart Specification enables models to learn the generative logic of charts with significantly improved data efficiency, reducing reliance on massive-scale supervision.
Moreover, our experiments show that spec-aligned reinforcement learning substantially enhances structural fidelity and logical coherence, successfully resolving complex data dependencies that standard supervised baselines fail to capture.

The main contributions of this work are summarized as follows:
\begin{itemize}
    \item We identify the cross-domain supervision limitation inherent in direct chart-to-code learning and propose Chart Specification as a structure-aware intermediate representation that bridges visual perception and code execution.

    \item We leverage this specification to construct ChartStruct, a dataset balanced by fine-grained structural categories, and design the Spec-Align Reward. This reward mechanism provides dense, outcome-oriented feedback, enabling effective reinforcement learning optimization.
    
    \item We achieve new state-of-the-art results across three public benchmarks with markedly superior data efficiency. Our method surpasses prior approaches on complex metrics using only 3K training samples and sets new performance records at 4K, successfully disentangling complex data dependencies where baselines fail.

\end{itemize}


\section{Related Work}
\subsection{Chart-to-Code Generation} Chart-to-code generation, aiming to reconstruct executable plotting code from chart images, has evolved from early neural translation to large-scale benchmarking. Pioneering works like Pix2Code \cite{beltramelli2017pix2code} and Chart2Code \cite{zhang2025chart2code} treated this as an image-to-text problem, yet they struggled with evaluation, as standard metrics like BLEU fail to capture the logical validity of code. Subsequent studies, including Plot2Code \cite{wu2024plot2code} and DePlot \cite{liu-etal-2023-deplot}, introduced larger datasets to improve generalization across plotting libraries. More recently, ChartCoder \cite{zhao2025chartcoder} scaled up the training data to 160k samples, attempting to enhance executability through massive supervised fine-tuning.

Compared with these fully supervised approaches, which often treat code as flat sequences and rely on scaling to cover corner cases, our work introduces Chart Specification as a structural intermediate representation. Instead of merely minimizing textual loss, we focus on aligning the topological structure of the generated code with the visual intent, enabling more data-efficient learning.

\subsection{Visual Reasoning for Charts} 

Recent advances in Vision-Language Models (VLMs) have significantly advanced chart reasoning tasks. Early efforts focused on fine-tuning large-scale pre-trained models for tasks like Chart Question Answering (ChartQA). For instance, \cite{xia2024chartx} apply supervised fine-tuning to extract relevant chart components for QA, while \cite{akhtar-etal-2023-reading} extends BERT-based architectures with chart features to enhance performance. 
To handle more complex logic, Chain-of-Thought (CoT) methodologies have been widely adopted. Approaches such as Multimodal CoT \cite{Wang2025MultimodalCoT} and CoT-VLA \cite{Zhao2025CoTVLA} break down reasoning into intermediate steps to improve decision-making. 
Meanwhile, capturing fine-grained structural relationships is a fundamental challenge shared across visual perception tasks, from 3D object detection \cite{zhang2025ascformer} to human-object interaction understanding \cite{han2026soft}. In the chart domain, layout-aware models like Pix2Struct \cite{Lee2022Pix2Struct} and MatCha \cite{Liu2022MatCha} pre-train on structural tasks (e.g. HTML parsing) to ground reasoning in visual layout, and Chart-CoF \cite{Li2025ChainOF} proposes a programmatic pipeline to synthesize fine-grained reasoning traces. 

Unlike these methods that rely largely on fixed templates or surface-level pattern matching for textual outputs, our approach aims at the generation of executable logic. We move beyond static QA by employing Reinforcement Learning to autonomously explore reasoning strategies, allowing the model to optimize its structural understanding without being constrained by the distribution of supervised reasoning templates.

\subsection{Reinforcement Learning with Verifiable Rewards (RLVR)} Reinforcement Learning (RL) has shifted from optimizing human preference (RLHF) \cite{ouyang2022training} toward leveraging verifiable rewards (RLVR) for objective tasks. Rather than relying on noisy human feedback or neural reward models as in DPO-style objectives \cite{rafailov2023dpo, liang2025chexpo}, recent work focuses on programmatic verifiers. Tülu 3 \cite{lambert2025tulu3} demonstrates that replacing neural rewards with simple correctness checks yields scalable improvements. In parallel, DeepSeekMath \cite{shao2024deepseekmath} and DeepSeek-R1 \cite{guo2025deepseekr1} adopt Group Relative Policy Optimization (GRPO) to train on binary rewards, unlocking strong chain-of-thought reasoning in math and coding tasks.
This verifiable paradigm has recently expanded to other structured domains, evidenced by R1-T1 \cite{he2025r1} for machine translation and R-Log \cite{liu2025rlog} for automated log analysis.
Building on this, recent algorithms further enhance the optimization stability: DAPO \cite{yu2025dapo} combines asymmetric clipping with dynamic sampling, GSPO \cite{zheng2025gspo} introduces sequence-level group optimization for MoE architectures, and DCPO \cite{yang2025dcpo} adopts token-wise dynamic clipping to maximize sample efficiency.

Our work differs from existing RLVR applications in the nature of the verification signal. While current methods predominantly focus on scalar correctness for math or single-step execution checks, we propose the Spec-Align Reward to address the ambiguity of visual generation. By quantifying structural fidelity rather than binary execution success, we extend the RLVR paradigm to the domain of visual reconstruction, closing the gap between scalar correctness and structured visual fidelity.

\section{Methodology}
\label{sec:method}

\begin{figure*}[t]
    \centering
    \includegraphics[width=\linewidth]{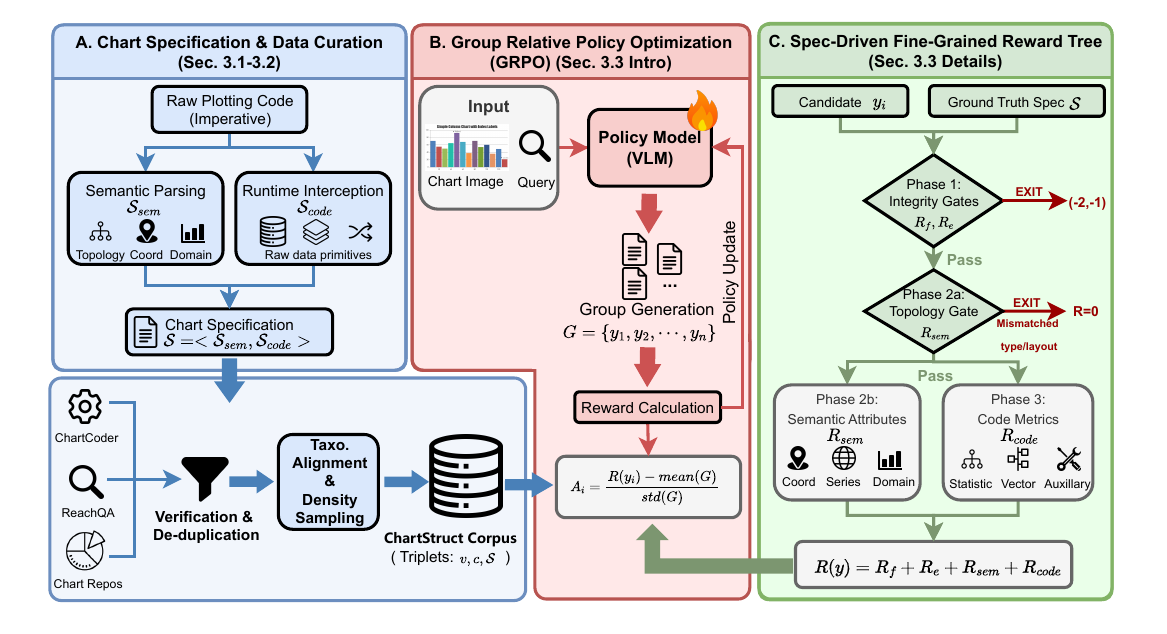} 
    \caption{The Overview of Our Framework. (A) Specification-Driven Data Curation: Adopting Chart Specification ($\mathcal{S}$) to extract semantic intent ($\mathcal{S}_{sem}$) and physical execution data ($\mathcal{S}_{code}$) from raw scripts, and guiding the curation of the ChartStruct corpus. (B) Group Relative Policy Optimization: The VLM policy is optimized using group-based advantage estimation. (C) Hierarchical Reward Tree: A fine-grained reward mechanism validates candidates through a staircase pipeline, checking Integrity (Phase 1) and Semantic Topology (Phase 2) before calculating precise Code Metrics (Phase 3).}
    \label{fig:pipeline}
\end{figure*}

In this section, we introduce our framework for chart-to-code generation.
The framework is centered on Chart Specification ($\mathcal{S}$), which serves as the interface across data curation, policy optimization, and fine-grained reward modeling.
As illustrated in Figure~\ref{fig:pipeline}, our method consists of three stages. 
We first define the schema of Chart Specification and an extraction pipeline that maps both reference and generated plotting code into this representation.
Then, based on the extracted specifications, we curate the ChartStruct training corpus by organizing and balancing samples according to fine-grained structural patterns.
Finally, we introduce Spec-Align Reward Modeling, which organizes supervision into a hierarchical Reward Tree, enabling dense and structured feedback for policy optimization.

\begin{figure*}[t]
    \centering
    \includegraphics[width=\textwidth]{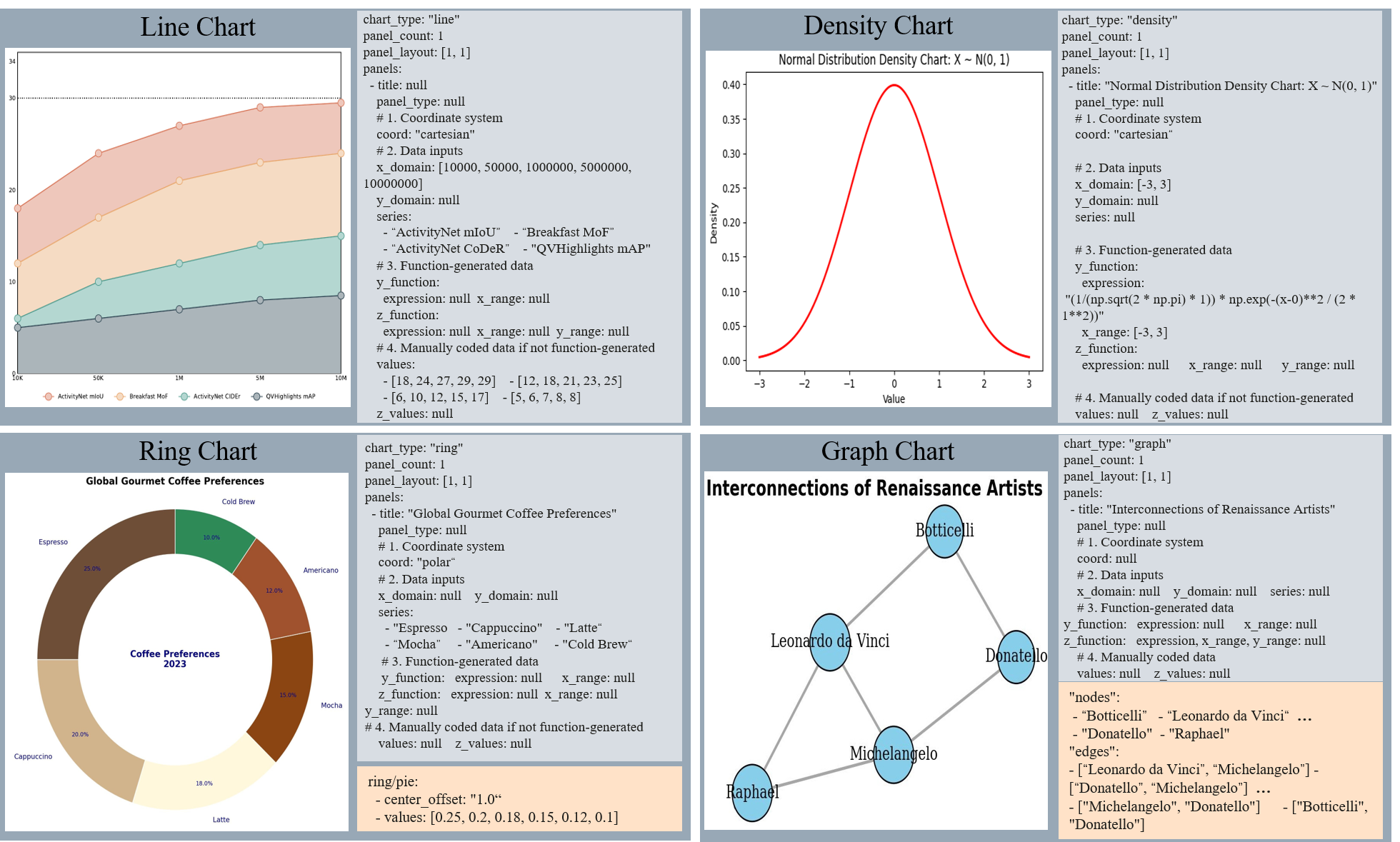} 
    \caption{Visualization of our Chart Specification ($\mathcal{S}$) across four distinct chart types. The grey panels represent the $\mathcal{S}_{sem}$, capturing declarative intents like topology and data domains. The \textbf{orange panels} (bottom-left and bottom-right) illustrate the $\mathcal{S}_{code}$, which uses runtime interception to capture implicit data, such as calculated wedge ratios in Ring charts or node-edge relationships in Network graphs.}
    \label{fig:spec_cases}
\end{figure*}

\subsection{Chart Specification: Schema Design and Validation}\label{sec:chartspec}

\textbf{Rationale and Schema Design.}
The design of Chart Specification ($\mathcal{S}$) draws on the decomposition principle of \textit{Grammar of Graphics} (GoG)~\cite{wilkinson2011grammar} and declarative grammars such as \textit{Vega-Lite}~\cite{satyanarayan2017vegalite}, which factor visualizations into orthogonal components that include data abstraction, geometric marks, coordinate systems, scales, and view composition. 
Inspired by this principle, we derive a \emph{task-oriented structural representation} tailored for chart code verification and reward computation, scoped to static, single-frame charts produced by imperative plotting libraries.  We adopt a hybrid two-layer design rather than a single declarative grammar, jointly covering the 20 chart families considered in this work, including those whose structure is awkward to express in purely declarative form, such as network graphs, vector fields, and treemaps.
 
Formally, we define $\mathcal{S}$ as a hybrid architecture:
\begin{equation}
    \mathcal{S} = \langle \mathcal{S}_{sem}, \mathcal{S}_{code} \rangle,
\end{equation}
where $\mathcal{S}_{sem}$ captures the semantic intent following the GoG decomposition and $\mathcal{S}_{code}$ grounds numerical and structural facts through execution. Although a single chart image may admit multiple imperative implementations, each instance in our corpus is associated with a fixed reference script; the specification abstracts away imperative syntax while preserving semantic intent, enabling consistent structural comparison across implementations. Table~\ref{tab:gog_mapping} details how the schema fields are organized along the dimensions of the GoG decomposition.

\begin{table}[t]
\centering
\caption{Chart Specification schema and its relation to the GoG decomposition.}
\label{tab:gog_mapping}
\small
\setlength{\tabcolsep}{4pt}
\begin{tabular}{@{}lcp{3.5cm}@{}}
\toprule
\textbf{GoG Dimension} & \textbf{Spec Layer} & \textbf{Schema Fields} \\
\midrule
Geometry / Mark       & $\mathcal{S}_{sem}$ & \texttt{chart\_type} \\
Facet / View Comp.    & $\mathcal{S}_{sem}$ & \texttt{panel\_count, panel\_layout} \\
Coordinate            & $\mathcal{S}_{sem}$ & \texttt{coord} \\
Scale                 & $\mathcal{S}_{sem}$ & \texttt{x\_domain, y\_domain} \\
Data / Encoding       & $\mathcal{S}_{sem}$ & \texttt{series, values, z\_values} \\
Transform             & $\mathcal{S}_{sem}$ & \texttt{y\_function, z\_function} \\
\midrule
Statistics            & $\mathcal{S}_{code}$ & \texttt{values, nodes, edges, \ldots, coords} \\
Auxiliary Render   & $\mathcal{S}_{code}$ & \texttt{positions, color} \\
\bottomrule
\end{tabular}
\end{table}

\textbf{Mapping Procedural Scripts to Semantic Intent ($\mathcal{S}_{sem}$).}
As illustrated in Figure~\ref{fig:spec_cases}, $\mathcal{S}_{sem}$ (grey panels) normalizes chart attributes into four structural dimensions:
(1) \textit{Global Topology}, capturing chart type, panel count and layout.  
(2) \textit{Coordinate Systems}, identifying spatial projections such as cartesian, polar, or 3d.  
(3) \textit{Data Domains}, defining axis ranges (x/y\_domain) and categorical series labels.  
(4) \textit{Analytic Representations}, abstracting data transformations into explicit functional forms.  

For instance, the Density Chart in Figure~\ref{fig:spec_cases} is specified by the extracted expression \texttt{np.exp(-(x-0)**2...)}, preserving analytic semantics beyond discrete samples. Notably, $\mathcal{S}_{sem}$ is designed to be human-verifiable by construction, with each field corresponding to a directly observable semantic attribute implied by the reference plotting code.  
In practice, $\mathcal{S}_{sem}$ fields are extracted from plotting scripts using a large language model Qwen3-32B with a carefully designed prompt, with the exception of \textit{panel\_count} and \textit{panel\_layout}.
We further conduct manual verification on a curated set of 200 instances drawn from the candidate training corpus, covering all 20 chart types in our taxonomy (10 samples per type).
For each chart type, the selected instances are visually diverse, exhibiting substantial structural variations in chart layouts and visual encodings.
This manual inspection confirms that the extracted $\mathcal{S}_{sem}$ are stable and semantically consistent with the reference code across diverse chart structures.

\textbf{Grounding Numeric via Runtime Interception ($\mathcal{S}_{code}$).}
While $\mathcal{S}_{sem}$ captures high-level intent, our empirical analysis reveals that purely text-based extraction struggles when chart data is implicitly computed or transformed during runtime. 
To address this limitation, we introduce $\mathcal{S}_{code}$ via runtime interception. Following script execution, we traverse the core plotting object hierarchy (e.g., \textit{Figure}, \textit{Axes}, and underlying \textit{Artist} objects) to capture the exact data primitives materialized by the rendering engine. Since most plotting frameworks expose analogous post-execution object layers, such as Plotly's \textit{Trace} hierarchy, this extraction paradigm could extend beyond \textit{matplotlib} with a framework-specific adapter to the schema.
As illustrated by the orange panels in Figure~\ref{fig:spec_cases}, $\mathcal{S}_{code}$ enables recovering implicit quantities such as wedge proportions in Ring charts (e.g., \texttt{values: [0.25, 0.2...]}) and explicit materialization of node-edge topology in Graph charts.

\begin{table}[t]
    \centering
    \caption{\textbf{The data distribution of a 4K instantiation for the \textsc{ChartStruct} corpus.}
    We define a target sample count per structural signature ($N_{target}$) based on structural complexity
    ($\rho \approx 90/72/54$ for Tier 1/2/3).}
    \label{tab:data_dist}
    \small
    \renewcommand{\arraystretch}{0.90}
    \setlength{\tabcolsep}{4pt}
    \resizebox{\columnwidth}{!}{%
        \begin{tabular}{llcccr}
            \toprule
            \textbf{Family} &
            \textbf{Source Mapping} &
            \textbf{Signature} &
            \textbf{Candidate} &
            \textbf{Count} &
            \textbf{Ratio} \\
            \cmidrule(lr){3-4}
            & & \textbf{Num} & \textbf{Pool} & & \\
            \midrule
            \rowcolor{gray!10} \multicolumn{6}{l}{\textit{\textbf{Tier 1: High Complexity (Target $\rho = 90$)}}} \\
            \texttt{mix} & \{combination, inset\} & 7 & 5237 & \textbf{630} & 15.8\% \\
            \texttt{3d} & \{3d\} & 4 & 5864 & \textbf{360} & 9.0\% \\
            \texttt{multi\_axes} & \{multi\_axes\} & 3 & 3279 & \textbf{270} & 6.8\% \\
            \texttt{radar} & \{radar\} & 2 & 4216 & \textbf{180} & 4.5\% \\
            \texttt{rose} & \{rose\} & 2 & 2827 & \textbf{180} & 4.5\% \\
            \texttt{contour} & \{line\}  & 2 & 641 & \textbf{180} & 4.5\% \\
            \texttt{quiver} & \{quiver\} & 2 & 2689 & \textbf{180} & 4.5\% \\
            \midrule
            \rowcolor{gray!10} \multicolumn{6}{l}{\textit{\textbf{Tier 2: Standard Scientific (Target $\rho = 72$)}}} \\
            \texttt{boxplot} & \{boxplot\} & 3 & 4873 & \textbf{216} & 5.4\% \\
            \texttt{pie} & \{pie\} & 2 & 5631 & \textbf{144} & 3.6\% \\
            \texttt{heatmap} & \{heatmap\} & 2 & 4264 & \textbf{144} & 3.6\% \\
            \texttt{error} & \{error bar\} & 2 & 1937 & \textbf{144} & 3.6\% \\
            \texttt{ring} & \{ring\} & 2 & 3268 & \textbf{144} & 3.6\% \\
            \texttt{violin} & \{violin\} & 1 & 3619 & \textbf{72} & 1.8\% \\
            \texttt{treemap} & \{treemap\} & 1 & 2987 & \textbf{72} & 1.8\% \\
            \midrule
            \rowcolor{gray!10} \multicolumn{6}{l}{\textit{\textbf{Tier 3: Basic Primitives (Target $\rho = 54$)}}} \\
            \texttt{bar} & \{bar, bar\_num\} & 7 & 15672 & \textbf{378} & 9.5\% \\
            \texttt{line} & \{line, area\} & 6 & 12846 & \textbf{324} & 8.1\% \\
            \texttt{scatter} & \{scatter, bubble\} & 4 & 6813 & \textbf{216} & 5.4\% \\
            \texttt{graph} & \{graph\} & 1 & 742 & \textbf{54} & 1.4\% \\
            \texttt{histogram} & \{histogram\} & 1 & 1217 & \textbf{54} & 1.4\% \\
            \texttt{density} & \{density\} & 1 & 681 & \textbf{54} & 1.4\% \\
            \midrule
            \textbf{Total} &  & \textbf{55} & 89303 & \textbf{3996} & \textbf{100\%} \\
            \bottomrule
        \end{tabular}%
    }
\end{table}

\subsection{Curation of the ChartStruct Training Corpus}\label{sec:data}

We propose ChartStruct, a spec-driven data curation pipeline for constructing
structurally balanced chart-to-code corpora from large, heterogeneous plotting code repositories.
Using this pipeline, we instantiate a 4K-scale corpus as the default training set for our main experiments, consisting of 3996 chart triplets $\mathcal{D} = \{(v, c, \mathcal{S})\}$, where $v$ is the chart image, $c$ the executable code, and $\mathcal{S}$ the corresponding Chart Specification.
The same pipeline naturally supports smaller instantiations by adjusting the target density of structural fingerprints; in particular, we construct a 3K-scale variant (3008 samples) that is used for data efficiency analysis and ablation studies.

\textbf{Taxonomy Alignment and Verification.}
The ChartStruct curation pipeline begins by aligning raw chart data into our unified taxonomy to ensure consistency with our subsequent data curation pipeline.
We curate our corpus primarily from ChartCoder~\cite{zhao2025chartcoder} and consolidate its 27 chart types into 20 canonical families by merging morphologically similar variants (e.g., area into line, bubble into scatter), while retaining structurally distinctive cases such as contour as separate canonical families when they are identified during our curation process.
The resulting per-sample taxonomy label is used directly as the ground-truth \textit{chart\_type} field in the Chart Specification to ensure annotation reliability.

Candidate samples then undergo a two-stage verification process.
The first stage re-executes all plotting code under a standardized environment, filtering out samples with latent runtime failures not caught by the original repository and reducing the pool to approximately 115K samples.
The second stage verifies structural integrity.
Although many scripts survive runtime execution, their source code often exhibits structural deficiencies inherited from large-scale template-driven code synthesis. 
These include degenerate plotting logic that merely varies surface-level parameters, or scripts whose plotting objects inherently lack essential attributes upon runtime traversal, making their underlying data-to-visual mappings unreliable or unrecoverable as training supervision.
This verification stage filters the raw repository into a Spec-valid candidate pool of around 89k samples, which serves for the next topological balancing stage.

\textbf{Structural Topology Tagging.}
With a naive inspection of the verified candidate pool, we observe a pronounced topological imbalance inherited from the raw data source. For example, in the verified ChartCoder pool, certain chart families (e.g., \texttt{bar}) contain over 10,000 valid samples, yet the majority differ only in surface-level numeric values while sharing identical plotting logic, obscuring the long tail of structurally
distinct chart configurations.
To address this issue, ChartStruct assigns high-level \emph{structural signatures} derived from the Chart Specification to each sample, characterizing a chart’s underlying topology beyond surface-level content.
We define a structural signature $\mathcal{S}_{struct}$ as a tuple of three core dimensions:
(1) \textit{Coordinate Space} (Cartesian, Polar, or 3D);
(2) \textit{Data Mode} (Explicit Values, Analytic Functions, or Matrix Grids); and
(3) \textit{Composition Topology} (Single-series, Grouped/Stacked Multi-series, or Subplots).
Enumerating these structural signatures $\mathcal{S}_{struct}$ across the verified pool
yields 55 distinct categories spanning the 20 chart families.

\textbf{Complexity-Adaptive Sampling.}
Following the above steps, we obtain a signature-annotated candidate pool. However, these structural signatures exhibit vastly different learning curves. Prior studies~\cite{han2023chartllama, zhao2025chartcoder} indicate that models struggle significantly more with complex layouts, such as 3D and nested subplots, due to the intricate spatial logic required. 
To address this, ChartStruct adopts a difficulty-aware sampling strategy, which allocates sampling density according to structural learning difficulty.
Constrained by a target budget of 4,000 samples for training efficiency,
we calibrate the target density $\rho$ to prioritize structural difficulty over raw frequency:
we allocate a higher quota ($\rho \approx 90$) to Tier~1 structures to guarantee sufficient coverage
of their complex configuration spaces, while limiting basic Tier~3 families to a minimal saturation point
($\rho \approx 54$) to prevent overfitting.
The detailed statistics of the resulting data distribution are reported in Table~\ref{tab:data_dist}.

\subsection{Spec-Align Fine-Grained Reward Modeling}
\label{sec::reward}

Our reward system leverages the Chart Specification ($\mathcal{S}$) to provide fine-grained and structure-aware feedback. This mechanism is implemented within the Group Relative Policy Optimization (GRPO) framework~\cite{shao2024deepseekmath, guo2025deepseekr1}, which optimizes the policy by comparing the relative quality of multiple responses sampled from the same prompt. 
To support advantage estimation in GRPO, we design a multi-stage reward mechanism that progressively differentiates superior and inferior generated code.
For a group of completions $G = \{y_1, y_2, \dots, y_n\}$ generated from the same prompt, the advantage $A_i$ for the $i$-th completion $y_i$ is calculated as:

\begin{equation} A_i = \frac{R(y_i) - \text{mean}(G)}{\text{std}(G)} \end{equation}

Here, $\mathrm{mean}(G)$ and $\mathrm{std}(G)$ denote the mean and standard deviation of the rewards within the group. The total reward $R(y)$ is a composite score formulated as:

\begin{equation} R(y) = R_f + R_e + R_{sem} + R_{code} \end{equation}

In this formulation, $R_f$, $R_e$, $R_{sem}$, and $R_{code}$ represent the Format, Execution, Semantic, and Code rewards, respectively. 
As illustrated by the Reward Tree in Figure~\ref{fig:pipeline} and detailed in Table~\ref{tab:reward_design}, these components are applied in a gated, staircase-like manner, where advanced structural rewards are only accessible after foundational constraints are satisfied. 
Importantly, reward components are conditionally instantiated based on chart semantics, rather than being uniformly applied across all chart types.

\paragraph{Initial Verification Phase.} 
The first stage of the sequence establishes the basic viability of the output. Format Reward ($R_f$) enforces the "think-then-answer" structure by verifying reasoning tags, penalizing any attempt to skip the step-by-step reasoning phase with a score of $-2$. The Execution Reward ($R_e$) subsequently assesses the runnability of the generated code. We assign a positive score of $0.5$ for successful compilation and a penalty of $-1$ for syntax errors. 
This asymmetric design treats execution success as a filtering signal during relative ranking, distinguishing functional candidates without allowing execution to dominate downstream structural rewards.

\paragraph{Hierarchical Semantic Refinement ($R_{sem}$).} 
The semantic-spec reward $R_{sem}$ targets chart families whose visual intent is primarily expressed through explicit data values or analytic transformations (e.g., bar, line, scatter, heatmap). It is evaluated through a hierarchical gated process.

(1) Topological Gate: The system first verifies global chart topology, including chart type, panel layout, and panel count, using a binary match criterion (Table~\ref{tab:reward_design}, Topology Gate). If this check fails, the computation of $R_{sem}$ is terminated immediately with an early exit, yielding reward score $0$ . This hard gating mechanism prevents noisy gradients from fine-grained attributes when the overall chart structure is incorrect, encouraging the model to master global organization before local details.

(2) Detailed Attribute Alignment: After passing the gate, $R_{sem}$ accumulates normalized scores in $[0,1]$ for panel-level attributes, including coordinate systems (Coord), axis or categorical domains (Domain), legend and grouping consistency (Series), and data values or analytic expressions (Data / Func).

Notably, the Data/Function component is instantiated only when raw values or analytic expressions are explicitly defined in the chart specification. For chart families such as boxplots or violins, whose semantics are conveyed through derived statistics, this component is intentionally left Null.

\begin{table*}[t]
\centering
\caption{Hierarchical reward components and fine-grained scoring criteria used for
Spec-Align reward modeling. The reward is organized into integrity checks, semantic
structure alignment ($R_{sem}$), and code-level fidelity ($R_{code}$).}
\label{tab:reward_design}
\footnotesize
\renewcommand{\arraystretch}{1.2}
\setlength{\tabcolsep}{5pt}

\begin{tabularx}{\textwidth}{p{3.1cm} p{2.2cm} X c}
\toprule
\textbf{Component} & \textbf{Metric} & \textbf{Description} & \textbf{Score ($R$)} \\
\midrule

\multicolumn{4}{l}{\textbf{Integrity Checks}} \\
\midrule
Format ($R_f$)
& Binary Match
& Think-then-Answer structure verification.
& $\{0, -2\}$ \\

Execution ($R_e$)
& Compilation
& Compilation check in sandbox.
& $\{0.5, -1\}$ \\

\midrule
\multicolumn{4}{l}{\textbf{Semantic-Spec Reward ($R_{sem}$)}} \\
\midrule

Topology Gate
& Binary Match
& Global chart structure consistency: chart type, panel layout, and panel count.
& Pass / $\{3\}$ \\

Coord
& Match
& Classification of coordinate system (e.g., Cartesian, polar, 3D).
& $[0,1]$ \\

Domain
& Range IoU
& Overlap of axis ranges or categorical domains.
& $[0,1]$ \\

Series
& Jaccard
& Consistency of legend labels and grouping sets.
& $[0,1]$ \\

Data / Func
& Edit Dist. / MSE
& Formula string matching or numeric sample error.
& $[0,1]$ \\

\midrule
\multicolumn{4}{l}{\textbf{Code-Spec Reward ($R_{code}$)}} \\
\midrule

Statistical
& $L_2$ Distance
& Boxplot quartiles or violin envelope statistics.
& $[0,1]$ \\

Relational
& F1-Score
& Treemap area ratios or graph node--edge relations.
& $[0,1]$ \\

Vector
& Cosine Sim.
& Vector field direction and magnitude alignment.
& $[0,1]$ \\

Auxiliary 
& Edit Dist.
& Text annotations, spatial positions, event markers, or color consistency.
& $[0,1]$ \\

\bottomrule
\end{tabularx}
\end{table*}

\paragraph{Fine-Grained Numerical Calibration ($R_{code}$).} 
Complementing to $R_{sem}$, the code-spec reward $R_{code}$ provides implementation-level feedback for chart families whose semantics are not fully captured by value- or function-level alignment but instead depend on statistical, relational, or geometric properties. Its components are selectively activated by chart family, avoiding conflicting or redundant supervision.
Specifically, $R_{code}$ instantiates four chart-family–specific metrics in Table~\ref{tab:reward_design}, computed from the runtime code-spec schema.

(1) Statistical. For boxplots, we extract per-category statistics (min, q1, median, q3, max) from predicted and reference executions and compute an $L_2$ distance over the concatenated statistic vectors; for violins, we compute the $L_2$ distance between envelope summaries (e.g., min/median/max and/or sampled boundary points). The distance is converted to a bounded score in $[0,1]$ via a monotonic normalization.
(2) Relational. For treemaps, we compare normalized area ratios of leaf nodes and score them by F1 based on a tolerance-aware match between predicted and reference leaves; for graphs, we compute F1 over recovered edge sets (and node/label sets when available) under canonicalized node identities.
(3) Vector. For quiver plots, we extract vector components $(\Delta x, \Delta y)$ at corresponding anchors and compute cosine similarity of directions (optionally weighted by magnitude); for contour plots, we compare recovered level sets (or level values when only levels are available) using set similarity.
(4) Auxiliary. For text and annotations, we compute edit distance on string content and Euclidean distance on normalized positions; for color, we compare color assignments under canonicalized palettes or label-aligned mappings. All metrics are normalized to $[0,1]$ for comparability.

These code-level rewards are mutually exclusive with semantic data-level components and are normalized to $[0,1]$ for numerical comparability.


\section{Experiments settings}
\subsection{Implementation Details}
We adopt Qwen2.5-VL-7B~\citep{wang2024qwen2vl} as the backbone model, chosen for its strong visual understanding capability and parameter efficiency~\citep{bai2025qwen2}.
Unless otherwise specified, all main experiments are trained on the 4K-scale ChartStruct instantiation described in Sec.~\ref{sec:data}, while a 3K-scale variant (3,008 samples) is used for ablation studies due to computational constraints.

We fine-tune the full parameters of the language model and the vision-language projector using Group Relative Policy Optimization (GRPO)~\citep{shao2024deepseekmath}, while keeping the visual encoder frozen to preserve pre-trained visual features.
Training is conducted for 3 epochs with a learning rate of $4\times10^{-7}$ and a global batch size of 32.
During exploration, we sample 16 rollouts per update to estimate group advantages, and set both prompt and response length limits to 3,000 tokens.
The KL divergence coefficient is fixed at 0.001 for training stability.
All experiments were run on 8 NVIDIA A100 (80GB) GPUs, completing in approximately 62 hours.

\subsection{Benchmarks and Evaluation Metrics}

We utilize three complementary benchmarks to evaluate the model's chart understanding and code generation capabilities across different domains and difficulty levels.

\paragraph{ChartMimic}~\citep{yang2024chartmimic} focuses on evaluating the ability of MLLMs to redraw publication-quality charts from ArXiv papers, focusing on the strict preservation of the original style and visual appearance. 
Following the official settings, we adopt the Direct Mimic task for evaluation on a test subset of 600 samples. 
We report the Execution Rate to assess code validity, alongside Low-Level metrics which compare specific code attributes (e.g., layout, color, text) against the ground truth to measure structural fidelity. In addition, High-Level scores are computed by GPT to assess the overall visual similarity.

\paragraph{Plot2Code}~\citep{wu2024plot2code} serves as a standard testbed for programmatic visualization skills, aiming to convert figures from Matplotlib galleries into executable code. 
We evaluate models on the Direct Asking task using the standard test set of 132 samples. 
Performance is assessed using three metrics: Pass Rate for executability, Text-Match for token-level similarity, and a model-based Rating (via GPT) to comprehensively judge the visual chart quality of the generated code.

\paragraph{ChartX}~\citep{xia2024chartx} constitutes a large-scale benchmark covering a diverse distribution of general chart types and visual styles.
To evaluate generalization capabilities beyond scientific figures, we select the Redrawing task and conduct evaluations on the test set of 1,152 samples.
We report the GPT-Score evaluated by GPT to measure the semantic consistency between the generated chart and the user instructions.

\subsection{Baseline Methods}

We compare our method against a comprehensive set of Multimodal LLMs (MLLMs). To ensure a fair and rigorous evaluation, we categorize these baselines into three distinct groups based on their accessibility and training domain:

\paragraph{Proprietary MLLMs.}
We include commercial models as high-performance references, namely GeminiProVision~\citep{team2023gemini}, Claude-3-Opus~\citep{anthropic2024claude}, GPT-4V~\citep{openai2023gpt4v}, GPT-4o~\citep{hurst2024gpt}, and GPT-5~\citep{openai2025gpt5} (reasoning\_effort=high). Results are taken from the official benchmark leaderboards where available, and obtained by us under the same protocol otherwise.

\paragraph{Open-Source MLLMs.}
We include a wide range of strong open-source generalist MLLMs. Specifically, we evaluate the InternVL series (8B, 26B, 76B)~\citep{zhang2024internlm}, Qwen2-VL-7B~\citep{wang2024qwen2vl}, LLaVA-Next~\citep{liu2024llavanext}, and MiniCPM-Llama3-V2.5~\citep{yao2024minicpm}. We also include recent open-source reasoning MLLMs, Qwen3-VL-8B~\citep{bai2025qwen3} and GLM-4.1V-9B~\citep{hong2025glm}, to provide a stronger test of our explicit Chart Specification design. Since our method is built on Qwen2.5-VL-7B~\citep{wang2024qwen2vl}, these baselines cover both standard models used in prior work and recent reasoning-oriented MLLMs.

\paragraph{Chart-Domain MLLMs.}
To contrast our structural reinforcement strategy with existing specialized approaches, we compare against models explicitly fine-tuned for chart-to-code generation tasks. This category includes TinyChart~\citep{zhang2024tinychart}, ChartVLM~\citep{xia2024chartx}, ChartCoder\citep{zhao2025chartcoder}, VisCodex~\citep{jiang2026viscodex} and JanusCoderV~\citep{sun2026januscoder}.

\begin{table*}[h]
\centering
\footnotesize
\caption{Main results on the ChartMimic benchmark comparing ChartSpec against leading proprietary and open-source models. The evaluation covers code executability, low-level visual fidelity (Text, Layout, Type, Color), and high-level visual assessment. The best performance among open-source MLLMs is indicated in \textbf{bold}.}
\vspace{1mm}
\begin{tabularx}{\textwidth}{ll*{8}{>{\centering\arraybackslash}X}}
\toprule
\multirow{2}{*}{\textbf{Models}} & \multirow{2}{*}{\textbf{Para.}} &
\multirow{2}{*}{\textbf{Exec. Rate}} &
\multicolumn{5}{c}{\textbf{Low-Level}} &
\multirow{2}{*}{\textbf{High-Level}} &
\multirow{2}{*}{\textbf{Overall}}\\
\cline{4-8}
& & & Text & Layout & Type & Color & Avg. \\
\midrule
Full score           & --     & 100   & 100   & 100   & 100   & 100   & 100   & 100   & 100   \\
\midrule
\multicolumn{10}{c}{\textit{Proprietary MLLMs}} \\
\midrule
\rowcolor{gray!15} GPT-5  & --     & 93.8  & 88.4  & 92.3  & 76.0  & 69.2  & 81.5  & 85.8  & 83.7  \\
\rowcolor{gray!15} GPT-4o       & --     & 93.2  & 81.5  & 89.8  & 77.3  & 67.2  & 79.0  & 83.5  & 81.2   \\
\rowcolor{gray!15} Claude-3-opus & --    & 83.3  & 66.8  & 83.1  & 49.9  & 42.1  & 60.5  & 60.1  & 60.3  \\
\rowcolor{gray!15} GeminiProVision & --  & 68.2  & 52.6  & 64.2  & 51.3  & 47.1  & 53.8  & 53.3  & 53.6  \\
\midrule
\multicolumn{10}{c}{\textit{Open-Source General MLLMs}} \\
\midrule
InternVL2-26B         & 26.0B  & 69.3  & 39.2  & 58.7  & 35.9  & 31.8  & 41.4  & 47.4  & 44.4  \\
InternVL2-Llama3-76B  & 76.0B  & 83.2  & 54.1  & 74.5  & 49.2  & 41.5  & 54.8  & 62.2  & 58.5  \\
InternVL3-8B          & 8B     & 70.6  & 51.7  & 61.8  & 47.2  & 45.2  & 51.5  & 58.7  & 55.1  \\
MiniCPM-Llama3-V-2.5   & 8.4B  & 80.3  & 30.7  & 49.6  & 38.6  & 27.6  & 36.6  & 42.1  & 39.4  \\
LLaVA-Next-Mistral-7B    & 7.6B & 59.7 & 14.0  & 31.1  & 19.8  & 17.8  & 20.7  & 21.3  & 21.0  \\
LLaVA-Next-Yi-34B       & 34.8B & 50.2 & 15.9  & 29.6  & 17.6  & 15.2  & 19.6  & 20.6  & 20.1  \\
DeepSeek-VL-7B           & 7B & 41.3 & 15.3  & 26.6  & 19.7  & 14.5  & 19.0  & 20.4  & 19.7  \\
Qwen2-VL-7B             & 7B  & 64.8 & 37.7  & 53.9  & 39.4  & 28.9  & 39.9  & 35.0  & 37.5  \\
Qwen2.5-VL-7B(backbone)  & 7B  & 69.2  & 48.6  & 59.8  & 55.6  & 44.4  & 52.1  & 63.2  & 57.7  \\
\midrule
\multicolumn{10}{c}{\textit{Open-Source reasoning MLLMs}} \\
\midrule
Qwen3-VL-8B-Thinking    & 8B  & 79.7  & 61.8  & 75.7  & 65.4  & 54.6  & 64.4  & 68.3  & 66.4  \\
GLM4.1-9B-Thinking      & 9B  & 79.3  & 68.7  & 76.5  & 66.5  & 58.9  & 67.7  & 70.6  & 69.2  \\
\midrule
\multicolumn{10}{c}{\textit{Open-Source Chart-Domain MLLMs}} \\
\midrule
ChartLlama            & 13B    & 57.5  & - & - & - & -  & 24.8  & 28.1  & 26.5  \\
ChartVLM-L            & 14.3B  & 83.2  & - & - & - & -  & 15.8  & 13.9  & 14.9  \\
ChartCoder            & 7B     & 91.4  & 67.2 & \textbf{95.0} & 78.5 & \textbf{69.0} & 77.4  & 74.0 & 75.7 \\
VisCodex              & 8B     & -  & - & - & - & -  & 74.8 & 74.1 & 74.5   \\
JanusCoderV           & 8B     & 83.5  & 67.7 & 81.8 & 78.6 & 57.5 & 72.4  & 73.7 & 73.1   \\
\midrule
ChartSpec(Ours) + 3k & 7B  & 90.2 & \textbf{73.7} & 90.7 & \textbf{83.7} & 65.7 & \textbf{78.4} & \textbf{81.4} & \textbf{79.9} \\
ChartSpec(Ours) + 4k & 7B  & \textbf{93.5} & \textbf{80.5} & 92.8 & 82.0 & 67.0 & \textbf{80.6} & \textbf{84.2}  & \textbf{82.4} \\
\bottomrule
\end{tabularx}
\label{tab:chartmimic}
\end{table*}

\begin{table}[t]
\centering
\scriptsize
\caption{Comparison of Model Performance on Plot2Code and ChartX.
Most results are taken directly from the official Plot2Code~\cite{wu2024plot2code} and ChartCoder~\cite{zhao2025chartcoder} evaluations. The best performance among open-source MLLMs is indicated in \textbf{bold}.}
\vspace{1mm}
\resizebox{\columnwidth}{!}{
\begin{tabular}{lcccc}
\toprule
\multirow{2}{*}{\textbf{Models}} &
\multicolumn{3}{c}{\textbf{Plot2Code}} &
\multicolumn{1}{c}{\textbf{ChartX}} \\
\cline{2-4} \cline{5-5}
& \textbf{Pass} & \textbf{Text} & \textbf{Rating} & \textbf{GPT} \\
\midrule
Full score & 100 & 100 & 10 & 5 \\
\midrule
\multicolumn{5}{c}{\textit{Proprietary MLLMs}} \\
\midrule
\rowcolor{gray!15} GPT-5               & 87.7 & 59.8 & 6.57 & 3.63  \\
\rowcolor{gray!15} GPT-4o              & 88.6 & 56.3 & 5.71 & -    \\
\rowcolor{gray!15} GPT-4V              & 84.1 & 57.7 & 6.48 & 2.63 \\
\rowcolor{gray!15} Claude-3-opus       & 84.1 & 57.5 & 4.37 & -    \\
\rowcolor{gray!15} GeminiProVision     & 68.2 & 53.6 & 5.06 & -    \\
\midrule
\multicolumn{5}{c}{\textit{Open-Source General MLLMs}} \\
\midrule
InternVL2-26B        & 81.3 & 43.1 & 3.42 & 1.70 \\
InternVL2-Llama3-76B & 85.6 & 46.6 & 3.89 & 1.74 \\
MiniCPM-Llama3-V2.5  & 76.3 & 37.3 & 2.61 & 1.66 \\
Qwen2-VL-7B          & 68.2 & 33.8 & 3.10 & 1.50 \\
Qwen2-VL-72B         & 72.0 & 53.4 & 4.26 & 1.69 \\
Qwen2.5-VL-7B        & 73.5 & 47.1 & 4.71 & 2.23 \\
\midrule
\multicolumn{5}{c}{\textit{Open-Source Reasoning MLLMs}} \\
\midrule
Qwen3-VL-8B-Thinking & 87.1 & 45.9 & 4.87 & 2.43 \\
GLM4.1-9B-Thinking   & 75.8 & 49.3 & 4.62 & 2.36 \\
\midrule
\multicolumn{5}{c}{\textit{Open-Source Chart-Domain MLLMs}} \\
\midrule
ChartLlama           & 58.4 & 40.3 & 2.32 & 0.94 \\
ChartVLM-L           & -    & -    & -    & 1.58 \\
JanusCoderV          & 82.6 & 49.6 & 4.83 & 2.31 \\
ChartCoder           & 87.9 & 54.5 & 4.50 & 2.09 \\
\midrule
ChartSpec (Ours) + 3k & 86.4 & 50.3 & 6.28 & 3.38 \\
ChartSpec (Ours) + 4k & \textbf{88.7} & \textbf{55.3} & \textbf{6.32} & \textbf{3.52} \\
\bottomrule
\end{tabular}
}
\label{tab:plot2code_main}
\end{table}

\section{Experimental Results}
\label{sec:results}

In this section, we present the empirical evaluation of ChartSpec. We first report comparative results against relevant baselines across the three benchmarks, followed by ablation studies that analyze the contributions of our data construction and reinforcement learning strategies.

\subsection{Evaluation on ChartMimic}
\label{subsec:chartmimic_results}

We first evaluate performance on ChartMimic, with quantitative results summarized in Table~\ref{tab:chartmimic}.
Our ChartSpec achieves leading performance among open-source MLLMs, including recent reasoning MLLMs and visual-programmatic baselines.
Even with the standard 3k training set, it attains an Overall Score of 79.9, outperforming recent open-source reasoning MLLMs such as Qwen3-VL-8B-Thinking (66.4) and GLM4.1-9B-Thinking (69.2), as well as strong chart-domain and visual-programmatic models like ChartCoder (75.7), VisCodex (74.5), and JanusCoderV (73.1); scaling to 4k further boosts the score to 82.4.
This demonstrates that our spec-driven alignment is highly data-efficient, achieving strong performance without massive pre-training or a larger backbone.

ChartSpec is also highly competitive with top-tier proprietary models, surpassing GPT-4o (81.2) and approaching the recent frontier model GPT-5 (83.7) with only a 7B backbone. This small gap is concentrated in perception-intensive dimensions, where GPT-5's massive-scale visual and OCR capabilities lead on Text (88.4) and Color (69.2). On the structural Type dimension, however, ChartSpec (82.0/83.7) exceeds GPT-5 (76.0), showing that precise structural supervision—rather than sheer model scale—drives fidelity on structure-critical aspects of the task.

Compared with ChartCoder, which excels in Layout (95.0) but degrades significantly in Text rendering (67.2), ChartSpec achieves a superior balance.
Our model maintains a high Layout score of 92.8 while simultaneously securing a Text score of 80.5, comparable to GPT-4o.
Finally, our method achieves the highest Execution Rate of 93.5\% among open-source models, confirming that the Spec-Align reward effectively minimizes syntax errors and runtime failures.

\subsection{Evaluation on Plot2Code and ChartX}
\label{sec:eval_plot2code_chartx}

To further assess the capability of ChartSpec on more unseen data distributions, we extended our evaluation to the Plot2Code and ChartX benchmarks. 

\paragraph{Performance on Plot2Code}
As shown in Table~\ref{tab:plot2code_main}, ChartSpec demonstrates exceptional robustness on the Plot2Code benchmark.
Our model achieves a Pass Rate of 88.7\%, the highest among all evaluated models, slightly surpassing top-tier proprietary models including GPT-4o (88.6\%) and GPT-5 (87.7\%), as well as Claude-3-Opus (84.1\%).
This result is particularly significant given the parameter disparity: ChartSpec (7B) outperforms the massive open-source generalist InternVL2-Llama3-76B (85.6\%) by a clear margin, and remains ahead of Qwen3-VL-8B-Thinking (87.1\%) and GLM4.1-9B-Thinking (75.8\%).
In terms of code quality, our model achieves a Text-Match score of 55.3, competitive with GPT-4o (56.3) and significantly higher than other 7B evaluations such as DeepSeek-VL (32.6) and Qwen2-VL (33.8), as well as JanusCoderV (49.6).
This confirms that our Spec-Align reward does not merely optimize for visual similarity but also ensures high semantic fidelity in the generated code.

\paragraph{Performance on ChartX}
The results on ChartX further validate the structural versatility of our approach.
ChartSpec achieves the best performance among open-source models with an overall GPT-score of 3.52, outperforming the previous best chart specialist ChartCoder (2.09) by a substantial margin while remaining competitive with the recent proprietary frontier model GPT-5 (3.63). It also surpasses recent strong baselines such as Qwen3-VL-8B-Thinking (2.43), GLM4.1-9B-Thinking (2.36), and JanusCoderV (2.31).
A type-wise breakdown (Table~\ref{tab:chartx_breakdown}) shows that while baselines often degrade on fine-grained or complex chart types, ChartSpec stays consistently strong: on geometrically complex categories such as \textit{Radar} and \textit{Bubble} charts, it scores 3.30 and 3.52 respectively, whereas GPT-4V only achieves 2.30 and 2.40.
This robustness further supports the effectiveness of our structural specification design on diverse chart topologies.

\begin{table*}[t]
\centering
\small
\setlength{\tabcolsep}{4pt}
\caption{\textbf{Type-wise accuracy for chart re-drawing on ChartX evaluated using GPT-score.}
All reported scores of baselines are obtained from the ChartX benchmark~\cite{xia2024chartvlm}.}
\label{tab:chartx_breakdown}
\vspace{1mm}
\resizebox{\textwidth}{!}{
\begin{tabular}{c ccccc ccccccccccccc c}
\toprule
\textbf{Models} &
\multicolumn{5}{c}{\textbf{General Chart Types}} &
\multicolumn{13}{c}{\textbf{Fine-grained Chart Types}} &
\textbf{Avg.} \\

\cmidrule(lr){2-6}
\cmidrule(lr){7-19}

& bar & bar\_num & line & line\_num & pie
& ring & box & hist & treemap & rose & area
& 3D-bar & bubble & multi & radar & heatmap & funnel & candle
& \\

\midrule
QWen-VL
& 0.89 & 0.60 & 0.80 & 1.30 & 1.25
& 1.10 & 0.80 & 0.80 & 0.80 & 1.10 & 0.60
& 1.10 & 0.90 & 0.60 & 0.50 & 0.50 & 0.70 & 0.70
& 0.86 \\

SPHINX-V2
& 1.00 & 1.75 & 1.60 & 1.65 & 1.80
& 0.50 & 0.40 & 1.60 & 0.60 & 1.10 & 0.20
& 0.50 & 0.40 & 0.20 & 0.00 & 0.50 & 0.30 & 0.20
& 0.96 \\

ChartLlama
& 1.16 & 1.05 & 0.90 & 1.15 & 1.80
& 0.70 & 0.80 & 1.00 & 1.00 & 0.70 & 0.70
& 1.10 & 0.70 & 0.40 & 0.50 & 1.00 & 0.70 & 0.30
& 0.94 \\

ChartAsst
& 0.95 & 1.35 & 0.00 & 0.60 & 0.30
& 0.00 & 1.50 & 2.40 & 0.60 & 1.70 & 1.80
& 0.60 & 0.60 & 1.20 & 0.00 & 0.00 & 2.10 & 0.00
& 0.82 \\

LLaVA-1.5
& 0.95 & 0.75 & 0.80 & 0.95 & 0.90
& 0.60 & 0.60 & 0.80 & 0.70 & 1.00 & 0.60
& 0.80 & 0.90 & 0.40 & 0.60 & 0.70 & 0.50 & 0.50
& 0.75 \\

GPT-4V
& 2.05 & 2.70 & 2.05 & 2.75 & 3.55
& 3.40 & 2.00 & 2.70 & 2.70 & 2.80 & 2.20
& 2.70 & 2.40 & 2.80 & 2.30 & 3.20 & \textbf{3.50} & 1.60
& 2.63 \\

ChartVLM-B
& 1.63 & 1.50 & 1.70 & 1.65 & 1.90
& 1.10 & 1.90 & 1.10 & 0.40 & 1.20 & 0.80
& 1.00 & 1.70 & 1.30 & 0.80 & 1.20 & 1.00 & 1.10
& 1.36 \\

ChartVLM-L
& 1.53 & 1.85 & 1.85 & 1.70 & 2.75
& 1.90 & 1.40 & 1.20 & 0.90 & 1.00 & 1.10
& 1.60 & 1.30 & 1.50 & 0.80 & 1.90 & 1.20 & 1.10
& 1.58 \\

\midrule
\textbf{ChartSpec + 3k}
& \textbf{3.50} & 3.48 & 3.84 & 3.83 & 3.79
& 3.64 & 3.27 & 3.59 & \textbf{2.38} & 1.99 & 3.79
& 2.74 & 3.34 & \textbf{3.41} & \textbf{3.41} & \textbf{3.57} & 3.25 & 2.45
& 3.38 \\

\textbf{ChartSpec + 4k}
& 3.43 & \textbf{3.52} & \textbf{4.11} & \textbf{4.11} & \textbf{4.22}
& \textbf{3.83} & \textbf{3.61} & \textbf{3.79} & 1.80 & \textbf{2.47} & \textbf{3.83}
& \textbf{3.17} & \textbf{3.52} & 3.35 & 3.30 & 3.49 & 3.28 & \textbf{2.76}
& \textbf{3.52} \\

\bottomrule
\end{tabular}
}
\end{table*}


\subsection{Ablation Study}

To systematically validate our core contributions and dissect the mechanisms behind the performance gains, we conduct ablation studies along four dimensions:
(1) We compare different data scales of SFT and Spec-Align RL to validate our claim that structural supervision offers superior sample efficiency compared to surface-level pattern imitation. (2) We analyze the positive effect of the Chart Specification in data construction pipeline by comparing it with random corpus curation. (3) We decompose the Spec-Align reward components across model scales to demonstrate how fine-grained signals outperform generic objectives. (4) Finally, we investigate the optimal utilization of the Spec-Align reward. Specifically, we compare direct policy alignment against reasoning-enhanced generation to determine which paradigm better maximizes the potential of our reward mechanism.

\subsubsection{Impact of Training Data Scale}
\label{sec:ablation_Data_Scaling}
We evaluate the impact of training data scale on model performance by comparing how SFT and Spec-Align RL respond to increasing data under the same data construction pipeline, and how these differences manifest in generalization behavior.

\begin{figure*}[t]
    \centering
    \includegraphics[width=\textwidth]{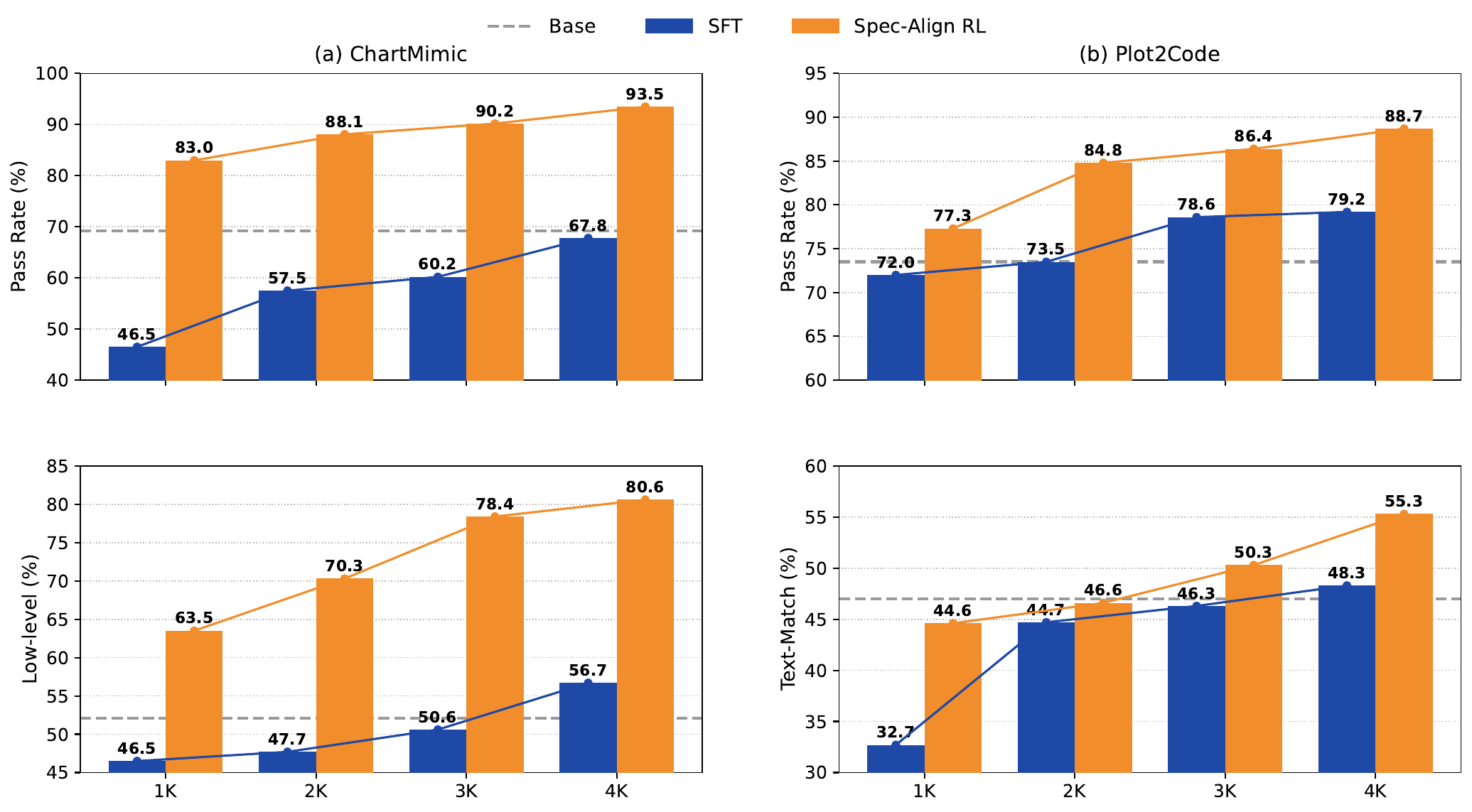}
    \caption{
    Impact of training data scale on chart-to-code performance.
    We compare SFT and Spec-Align RL across varying data sizes on ChartMimic (a) and Plot2Code (b), evaluated using Pass Rate, Low-level accuracy, and Text-Match.
    }
    \label{fig:data_scaling}
\end{figure*}

\paragraph{Sample Efficiency and Scaling Laws}
As illustrated in Figure~\ref{fig:data_scaling}, the two training paradigms exhibit fundamentally different scaling behaviors.
SFT suffers from an initial performance regression. At the 1K data regime, the Pass Rate on ChartMimic drops significantly to 46.5\%, falling well below the Base model's 69.2\%.
Even as the data scale increases to 4K, SFT only recovers to 67.8\%, struggling to surpass the original pre-training baseline.
This implies that standard SFT consumes substantial capacity re-learning low-level patterns, making it data-inefficient for this task.
In contrast, Spec-Align RL demonstrates exceptional sample efficiency.
With only 1K samples, it achieves an 83.0\% Pass Rate, outperforming the best SFT result by a margin of 15.2\%.
Performance improves rapidly at 3K (reaching 90.2\%) before showing signs of saturation at 4K.
This confirms that the structured reward signal enables the model to grasp generative rules much faster than surface-level supervision.

\begin{figure*}[t]
    \centering
    \includegraphics[width=\textwidth]{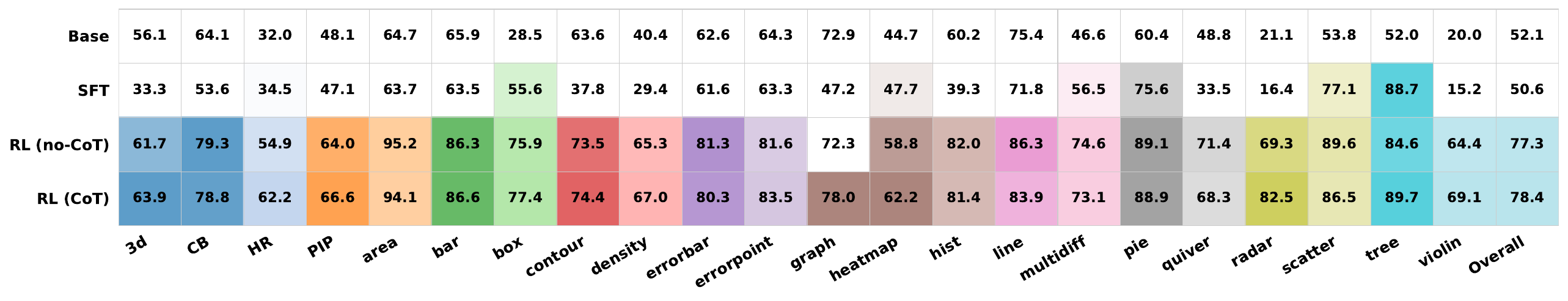}
    \caption{
    \textbf{Type-wise performance comparison at the 3K data scale on Chartmimic.} Here, low level average accuracy across different chart types. Columns correspond to chart categories, and rows denote different training settings:
    \textbf{Base} (Qwen2.5VL-7B-Instruct), \textbf{SFT} (Standard Fine-Tuning), \textbf{RL (no-CoT)} (Spec-Align RL without reasoning), and \textbf{RL (CoT)} (Spec-Align RL with reasoning).    
    The Overall column reports weighted averages over chart types.
    }
    \label{fig:heatmap_type}
\end{figure*}

\paragraph{Structural Generalization vs. Degradation}
To understand the source of this efficiency gap, we analyze the type-wise performance in Figure~\ref{fig:heatmap_type}.
The breakdown reveals that SFT is prone to structural degradation in complex chart types.
Although it maintains competence on standard formats, performance drops below the baseline on intricate geometries like \textit{Contour} (63.6 $\rightarrow$ 37.8) and \textit{Violin} plots (20.0 $\rightarrow$ 15.2).
This suggests that SFT tends to forget the coordinate mapping logic required for complex topologies.
Conversely, Spec-Align RL ensures robust generalization.
Not only reverses this degradation, it achieves dominant performance in these long-tail categories, raising \textit{Violin} scores to 69.1 and \textit{Graph} to 78.0.
This indicates that the Spec-Align reward encourages the internalization of generalized plotting logic rather than overfitting to simple, high-frequency patterns.

\subsubsection{Role of Chart Spec in Data Construction}
\label{sec:ablation_data}
We analyze the role of Chart Specification in data construction by comparing our spec-driven ChartStruct pipeline against a baseline trained on randomly curated data. Both strategies are applied to the 3K data scale and evaluated on Qwen2-VL and Qwen2.5-VL backbones.

As presented in Table \ref{tab:ablation_data}, the results highlight a critical distinction between code executability and semantic fidelity.
The model trained on ChartStruct demonstrates superior capability in grounding generated code to the underlying data.
Specifically, on the Plot2Code benchmark, the \textit{Text-Match} score significantly improves from 48.5 to 50.3 for Qwen2.5-VL-7B and from 37.7 to 41.7 for Qwen2-VL-7B.
A similar trend is observed on ChartMimic, where the \textit{Low-Level} metric consistently favors the Spec-driven models, exemplified by an increase from 76.6 to 78.4 on the Qwen2.5-VL backbone.
These metrics indicate that training on spec-verified data effectively forces the model to attend to precise coordinate mappings rather than hallucinating plausible but incorrect values.

\begin{table}[t]
    \centering
    \caption{Ablation study on data curation strategies. Comparison between Random Curation and our Spec-Driven ChartStruct.}
    \label{tab:ablation_data}
    \footnotesize
    \setlength{\tabcolsep}{4pt}
    \renewcommand{\arraystretch}{1.1}
    \begin{tabularx}{\columnwidth}{X c c c c}
        \toprule
        \multirow{2}{*}{\textbf{Model}} & \multicolumn{2}{c}{\textbf{ChartMimic}} & \multicolumn{2}{c}{\textbf{Plot2Code}} \\
        \cmidrule(lr){2-3} \cmidrule(lr){4-5}
        & \textbf{Pass} & \textbf{Low} & \textbf{Pass} & \textbf{Text} \\
        \midrule
        Qwen2-VL-7B (Random) & 88.2 & 72.9 & 88.3 & 37.7 \\
        \textbf{+ChartStruct (Ours)} & \textbf{91.7} & \textbf{73.8} & \textbf{92.4} & \textbf{41.7} \\
        \midrule
        Qwen2.5-VL-7B (Random) & 90.8 & 76.6 & 87.1 & 48.5 \\
        \textbf{+ChartStruct (Ours)} & 90.2 & \textbf{78.4} & 86.4 & \textbf{50.3} \\
        \bottomrule
    \end{tabularx}
\end{table}

This improvement in semantic precision accompanies an interesting observation regarding the Pass Rate, where minor fluctuations are observed on the Qwen2.5-VL backbone.
Specifically, the Spec-driven model achieves a Pass Rate of 90.2\% on ChartMimic and 86.4\% on Plot2Code, both of which are marginally lower than the Random baselines (90.8\% and 87.1\%, respectively).
We attribute this slight and consistent decrease to differences in data difficulty induced by spec-driven curation rather than reduced executability or training instability.
The Random baseline is more likely to be biased toward simpler, standard chart patterns that compile easily, whereas ChartStruct enforces a balanced inclusion of diverse and structurally complex topologies.
Importantly, this modest fluctuation contrasts with the substantial and consistent gains in Text-Match and Low-Level metrics, indicating that spec-driven curation shifts the model away from surface-level memorization toward structurally faithful and semantically grounded generation.


\begin{figure*}[t]
    \centering
    \includegraphics[width=\textwidth]{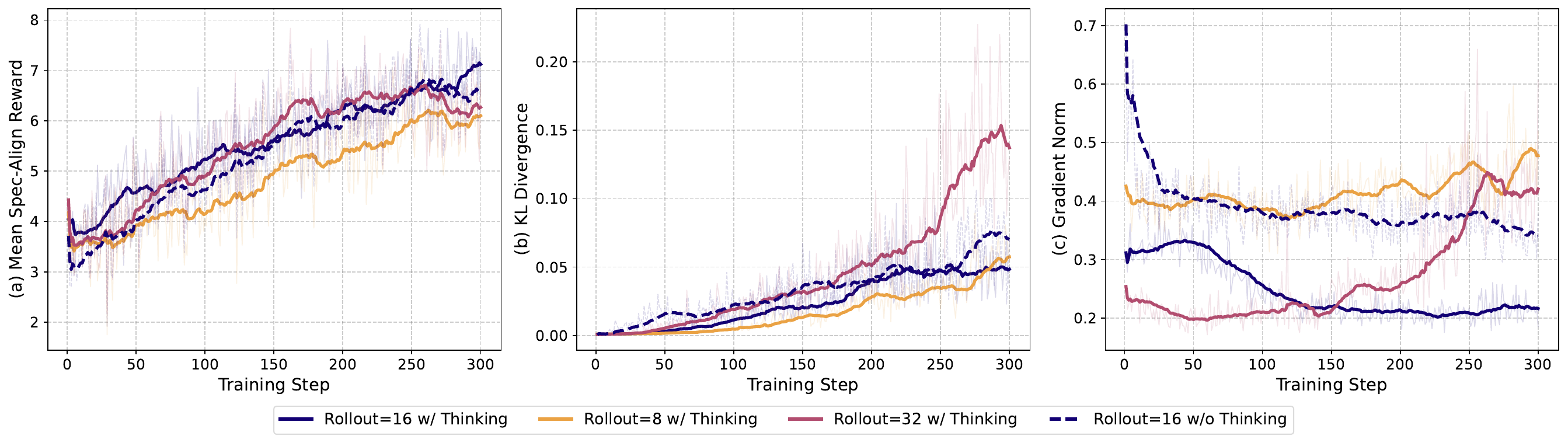}
    \caption{\textbf{Training dynamics under different rollout budgets and reasoning configurations.} (a) Mean Spec-Align reward, (b) KL divergence, and (c) gradient norm over training steps. The dashed line denotes training without the reasoning process (rollout=16).}
    \label{fig:training_dynamics}
\end{figure*}

\subsubsection{Efficacy and Decomposition of Reward Mechanisms}
\label{sec:ablation_reward}
We examine the reward mechanism from three perspectives: (i) the overall efficacy of the Spec-based RL paradigm across varying model scales, (ii) the individual contributions of semantic versus code-level specifications, and (iii) the training stability and topological robustness of the proposed reward-driven optimization.

\paragraph{Efficacy of Spec-Based Reward Mechanisms}
To demonstrate the effectiveness and broad generalizability of our approach, we conduct extensive experiments across multiple model families (Qwen2-VL, Qwen2.5-VL) and varying parameter scales (2B, 3B, 7B). We compare three settings: the vanilla Base model, Supervised Fine-Tuning (SFT), and our Spec-based RL (+Spec).

\begin{table}[t]
    \centering
    \caption{Ablation study of different training strategies across various model scales (+Spec denotes our method).}
    \label{tab:ablation_rl_main}
    \footnotesize
    \setlength{\tabcolsep}{2.5pt}
    \renewcommand{\arraystretch}{1.15}

    \begin{tabularx}{\columnwidth}{@{}Xcccc@{}}
        \toprule
        \textbf{Model} &
        \multicolumn{2}{c}{\textbf{ChartMimic}} &
        \multicolumn{2}{c}{\textbf{Plot2Code}} \\
        \cmidrule(lr){2-3} \cmidrule(lr){4-5}
        & \textbf{Pass} & \textbf{Low} & \textbf{Pass} & \textbf{Text} \\
        \midrule
        Q2.5-VL-7B & 69.2 & 52.1 & 73.5 & 47.1 \\
        Q2.5-VL-7B+SFT & 55.7 & 50.6 & 67.4 & 31.6 \\
        \textbf{Q2.5-VL-7B+Spec} & \textbf{90.2} & \textbf{78.4} & \textbf{86.4} & \textbf{50.3} \\
        \midrule
        Q2.5-VL-3B & 55.0 & 37.3 & 75.8 & 37.0 \\
        Q2.5-VL-3B+SFT & 48.0 & 36.2 & 57.6 & 35.7 \\
        \textbf{Q2.5-VL-3B+Spec} & \textbf{89.2} & \textbf{64.2} & \textbf{84.8} & \textbf{42.1} \\
        \midrule
        Q2-VL-7B & 64.8 & 39.9 & 68.2 & 33.8 \\
        Q2-VL-7B+SFT & 53.7 & 41.2 & 68.9 & 32.6 \\
        \textbf{Q2-VL-7B+Spec} & \textbf{91.7} & \textbf{73.8} & \textbf{92.4} & \textbf{41.7} \\
        \midrule
        Q2-VL-2B & 43.3 & 25.3 & 78.0 & 30.7 \\
        Q2-VL-2B+SFT & 37.0 & 25.8 & 59.8 & 23.9 \\
        \textbf{Q2-VL-2B+Spec} & \textbf{94.5} & \textbf{55.6} & \textbf{90.9} & \textbf{31.2} \\
        \bottomrule
    \end{tabularx}
\end{table}

The results in Table \ref{tab:ablation_rl_main} reveal a compelling insight: Supervised Fine-Tuning (SFT) alone is insufficient and often detrimental for this task. Across almost all evaluated models, applying SFT leads to a drop in Pass Rate compared to the Base model (e.g., Q2.5-VL-7B drops from 69.2\% to 55.7\% on ChartMimic). This phenomenon suggests that standard SFT may cause the model to overfit to specific text formats while losing the precise reasoning logic required for plotting. In sharp contrast, our Spec-based RL consistently reverses this trend and achieves state-of-the-art performance. The Q2.5-VL-7B model sees a massive jump to 90.2\% Pass Rate with our method. Furthermore, the improvements are highly scalable; even the smaller Q2-VL-2B model achieves a remarkable 94.5\% Pass Rate on ChartMimic with our framework, outperforming the base 7B models. This demonstrates that our reward mechanism provides a generalized solution for chart-to-code tasks, independent of the underlying backbone architecture.

\begin{table}[t]
    \centering
    \caption{Ablation study of reward configurations on Qwen2.5-VL-7B.}
    \label{tab:ablation_study}
    \footnotesize
    \setlength{\tabcolsep}{4pt}
    \renewcommand{\arraystretch}{1.15}
    \begin{tabularx}{\columnwidth}{@{}Xcccc@{}}
        \toprule
        \multirow{2}{*}{\textbf{Reward Configuration}} &
        \multicolumn{2}{c}{\textbf{ChartMimic}} &
        \multicolumn{2}{c}{\textbf{Plot2Code}} \\
        \cmidrule(lr){2-3} \cmidrule(lr){4-5}
        & \textbf{Pass} & \textbf{Low} & \textbf{Pass} & \textbf{Text} \\
        \midrule
        Qwen2.5-VL-7B (Base)          & 69.2 & 52.1 & 73.5 & 47.1 \\
        + Format \& Execution   & 97.3 & 61.2 & 96.2 & 36.8  \\
        \midrule
        + Semantic Spec                & 81.2 & 67.0 & 84.1 & 37.1 \\
        + Spec-Align (w/o Gate)         & 86.7  & 73.6  & 84.8  & 43.3  \\
        + Spec-Align (Full)             & \textbf{90.2} & \textbf{78.4} & \textbf{86.4} & \textbf{50.3} \\
        \bottomrule
    \end{tabularx}
\end{table}

\paragraph{Decomposition of Reward Components}
To further dissect the source of these gains, we analyze the specific contributions of each reward component. We use the Qwen2.5-VL-7B model as the baseline and incrementally introduce the reward components.

As illustrated in Table \ref{tab:ablation_study}, the baseline Qwen2.5-VL-7B model exhibits a notable deficit in structural precision. Adding Format \& Execution alone boosts Pass Rate substantially but yields limited fine-grained improvement and even degrades textual accuracy (Text: 47.1$\to$36.8), suggesting the model learns to produce executable code without faithfully reproducing chart content. The augmentation with Semantic Specifications primarily enhances the model's understanding of chart types and layouts. Incorporating Spec-Align without the Topology Gate (Row~4) further improves all metrics, confirming that Code-Spec signals provide additional value. However, a notable gap remains on fine-grained metrics (Low: 73.6 vs 78.4; Text: 43.3 vs 50.3), as structurally mismatched samples receive spurious fine-grained rewards that introduce optimization noise. The integration of full Spec-Align yields the most substantial enhancement across all metrics, validating that the Topology Gate ensures detailed rewards are computed only when global chart organization is correct. This underscores the role of explicit data descriptors in mitigating semantic hallucinations, ensuring that the generated code is not only visually similar but numerically precise.

\begin{table}[ht]
    \centering
    \caption{Hyperparameter robustness analysis.}
    \label{tab:robustness}
    \footnotesize
    \setlength{\tabcolsep}{4pt}
    \renewcommand{\arraystretch}{1.1}
    \begin{tabularx}{\columnwidth}{X c c c c}
        \toprule
        \multirow{2}{*}{\textbf{Setting}} &
        \multicolumn{2}{c}{\textbf{ChartMimic}} &
        \multicolumn{2}{c}{\textbf{Plot2Code}} \\
        \cmidrule(lr){2-3} \cmidrule(lr){4-5}
        & \textbf{Pass} & \textbf{Low} & \textbf{Pass} & \textbf{Text} \\
        \midrule
        Rollout = 8  & 88.8 & 76.9 & 84.1 & 42.5 \\
        Rollout = 16 (default) & 90.2 & 78.4 & 86.4 & 50.3 \\
        Rollout = 32 & 88.3 & 77.6 & 83.7 & 44.2 \\
        \bottomrule
    \end{tabularx}
\end{table}

\paragraph{Training Stability and Hyperparameter Robustness}
To verify the optimization stability and sensitivity of our GRPO training, we monitor key training diagnostics across different rollout budgets (8, 16, 32). As shown in Figure~\ref{fig:training_dynamics}(a), the mean Spec-Align reward increases steadily and converges across all three settings. The KL divergence remains bounded below 0.15 throughout training in Figure~\ref{fig:training_dynamics}(b), confirming the absence of policy drift. Table~\ref{tab:robustness} further shows that all rollout budgets achieve comparable structural fidelity, with ChartMimic Low-level scores varying within 2 percentage points (76.9--78.4). The primary difference lies in fine-grained text alignment: rollout=16 attains the highest Plot2Code Text-Match (50.3), outperforming both rollout=8 (42.5) and rollout=32 (44.2). We attribute this to the precision of advantage estimation in GRPO: smaller groups yield noisier advantage estimates due to limited samples, while excessively large groups amplify noisy normalized advantages as the within-group standard deviation shrinks during late training. Rollout=16 provides the most balanced trade-off and is adopted as the default.

\subsubsection{Impact of Reasoning Process}
\label{sec:reasoning_analysis}

We further investigate whether the reasoning process is truly necessary, or whether the Spec-Align reward alone can drive effective code generation without intermediate thinking steps.

\paragraph{Breaking the Performance Ceiling}
Table~\ref{tab:thinking_ablation} provides a detailed comparison of the two training paradigms. The model trained with Spec-driven RL but without reasoning steps already achieves a substantial improvement over the baseline, increasing the Pass Rate by 19.3\%. This proves that the reward signal itself is highly effective at teaching the model the general syntax of plotting code. However, this direct generation approach eventually hits a performance plateau, particularly on metrics demanding high precision. Incorporating the reasoning process allows the model to break this ceiling. This is most evident on the Plot2Code benchmark, where the reasoning model achieves a clear gain in Text-Match score (rising from 47.4 to 50.3). This improvement suggests that the intermediate reasoning tokens help the model accurately calibrate fine-grained details, such as specific color codes and legend positioning, which the direct model often fails to ground accurately.

\begin{table}[ht]
    \centering
    \caption{Ablation study on the impact of the reasoning process. \textit{w/o Thinking} refers to the model trained to generate code directly, while \textit{w/ Thinking} incorporates the intermediate reasoning chain.}
    \label{tab:thinking_ablation}
    \footnotesize
    \setlength{\tabcolsep}{3pt}
    \renewcommand{\arraystretch}{1.15}

    \begin{tabularx}{\columnwidth}{@{}Xcccc@{}}
        \toprule
        \multirow{2}{*}{\textbf{Model Variant}} &
        \multicolumn{2}{c}{\textbf{ChartMimic}} &
        \multicolumn{2}{c}{\textbf{Plot2Code}} \\
        \cmidrule(lr){2-3} \cmidrule(lr){4-5}
         & \textbf{Pass} & \textbf{Low} & \textbf{Pass} & \textbf{Text} \\
        \midrule
         Q2.5-VL-7B (Base)             & 69.2 & 52.1 & 73.5 & 47.1 \\
         \textit{RL w/o Thinking}    & 88.5 & 77.3 & 85.8 & 47.4 \\
         \textit{RL w/ Thinking}     & \textbf{90.2} & \textbf{78.4} & \textbf{86.4} & \textbf{50.3} \\
        \bottomrule
    \end{tabularx}
\end{table}

\paragraph{Handling Structural Complexity}
To further understand where reasoning is most beneficial, we analyze the performance gains across different chart categories in Figure~\ref{fig:heatmap_type}. The results show that the impact of reasoning is highly dependent on chart complexity. For standard chart types like Bar or Line plots, the performance gap between the two variants is negligible. This implies that direct visual-to-code mapping is sufficient for explicit and simple layouts. In contrast, the benefits of reasoning become dominant in structurally complex categories. We observe significant improvements in Tree and Violin plots, where the Pass Rate increases by 5.1\% and 4.7\% respectively. These findings confirm that while direct generation handles standard patterns well, the reasoning chain provides a necessary buffer to decompose intricate topological relations before implementation, effectively resolving ambiguities in complex geometries.

\begin{figure*}[t]
    \centering
    \includegraphics[width=\textwidth]{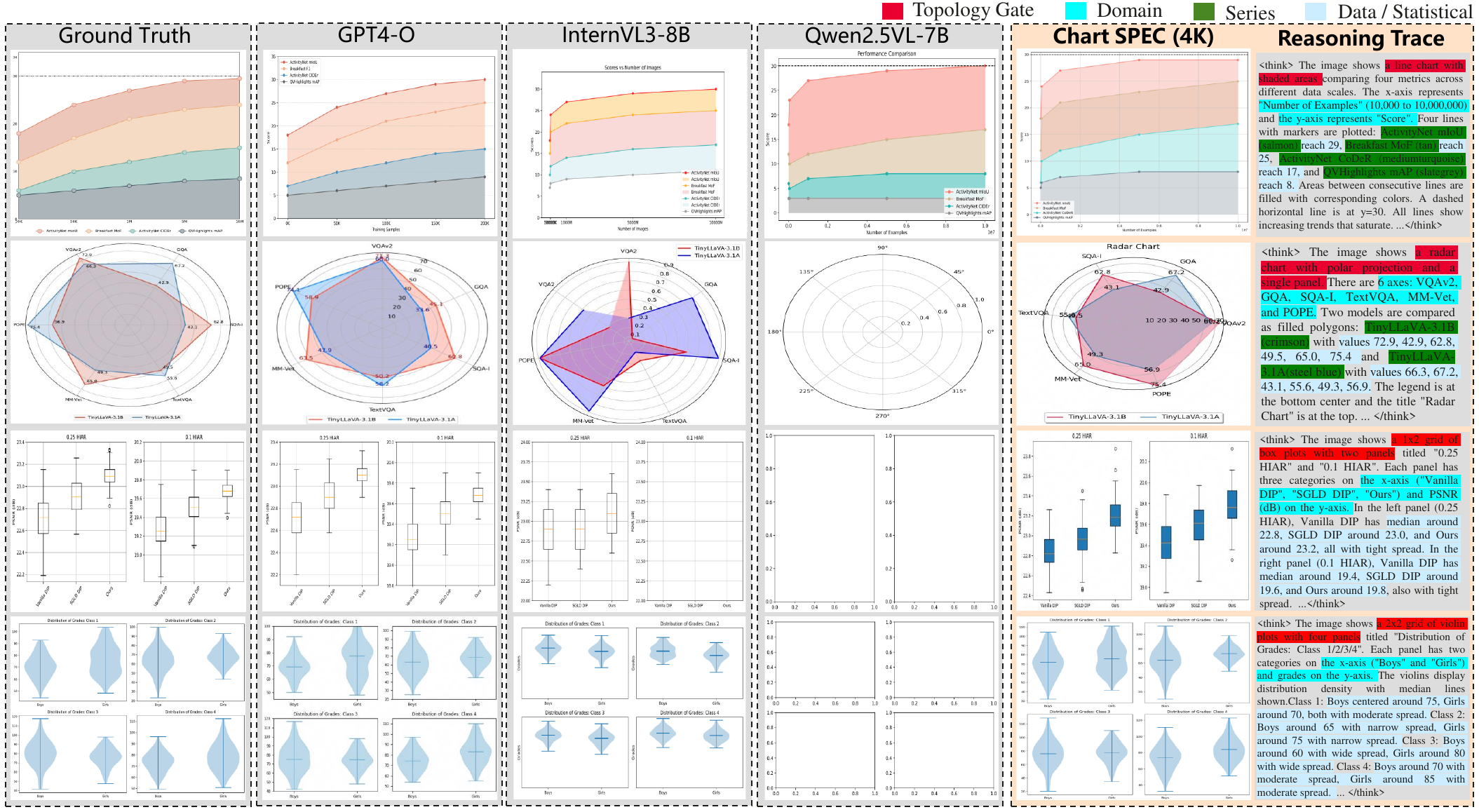} 
    \caption{Qualitative comparison and reasoning trace analysis on ChartMimic. Columns 1--5: Ground Truth, GPT-4o, InternVL3-8B, Qwen2.5-VL-7B, and ChartSpec 4K (Ours). Column 6: ChartSpec's reasoning trace before code generation, with color highlights indicating correspondence to our spec reward components.}
    \label{fig:qualitative_cases}
\end{figure*}

\paragraph{Interaction Between Reasoning and Reward Optimization.}
We attribute the above patterns to the role of reasoning in improving trajectory-level credit assignment under our Spec-Align reward. In direct code generation, the model must resolve chart topology, axis semantics, series identities, data values, and code realization within a single sequence, which entangles different sources of errors. The reasoning process factorizes this into a structured planning phase and a code realization phase. As shown in Figure~\ref{fig:qualitative_cases} (Column 6), the reasoning trace explicitly verbalizes reward-relevant factors, including topology, domain, series, and data/statistical properties, which correspond to the structural dimensions evaluated by the Spec-Align reward.

Although the Spec-Align reward is computed only from the final code, GRPO assigns the group-relative advantage to the entire generated trajectory, including the reasoning tokens. Reasoning traces that better decompose the chart structure are therefore implicitly reinforced through higher downstream rewards. For simple charts, structural decisions are few and can be resolved implicitly in code; for complex charts with many interacting factors, explicit decomposition becomes essential. This explains both the type-wise concentration of gains on complex charts (Figure~\ref{fig:heatmap_type}) and the training dynamics in Figure~\ref{fig:training_dynamics}: the \textit{w/ Thinking} variant reaches a higher final reward ((a), 7.0 vs 6.5) and exhibits smoother gradient norms from the outset ((c)).

\subsection{Analysis and Discussion}
\label{sec:analysis}

To intuitively evaluate generation quality, we visualize comparisons between \textit{ChartSpec} and representative baselines in Figure~\ref{fig:qualitative_cases}. Standard open-source VLMs (e.g., Qwen2.5, InternVL3) frequently fail on structurally complex charts: Qwen2.5-VL-7B produces an empty Radar skeleton with no data polygons (Row 2) and collapses the Violin y-axis to a 0–1 range (Row 4), while InternVL3 distorts the Radar shape beyond recognition. In contrast, ChartSpec produces faithful outputs across all four rows, rivaling or exceeding GPT-4o. On the Line chart (Row 1), our model correctly recovers the four trend lines, shaded regions, and axis labels. On the Radar chart (Row 2), ChartSpec closely matches the overall polygon geometry and relative axis magnitudes, while GPT-4o exhibits noticeable distortion in axis proportions. On the Violin chart (Row 4), ChartSpec recovers the correct grade ranges and produces distribution shapes closer to the ground truth than GPT-4o, reflecting the effect of the $R_{code}$ Statistical reward on envelope fidelity. On the Boxplot (Row 3), our model preserves the panel layout and axis semantics, though the rendered style differs from the ground truth, suggesting room for improvement in distinguishing visually similar chart families.

\section{Conclusion}
\label{sec:conclu}
In this paper, we introduced a verifiable reinforcement learning framework to bridge the gap between visual perception and code generation. By shifting from textual code imitation to Semantically Structured Grounding, our approach leverages Chart Specification to drive both high-quality data curation and the Spec-Align Reward, effectively enforcing rigorous plotting constraints and mitigating hallucinations. Extensive experiments demonstrate that our method achieves state-of-the-art results with exceptional data efficiency, surpassing competitive baselines using only 3K samples.

\section{Limitations}
\label{sec:limit}
Our evaluation is conducted entirely on matplotlib: the reference programs of all three benchmarks are written in matplotlib, and the fine-grained metrics of ChartMimic are computed over matplotlib runtime objects. We have therefore not verified empirically that Chart Specification transfers to other plotting frameworks, since a comparable evaluation would require new chart–code pairs and a reimplementation of these metrics over another framework's object model. A second limitation concerns scale: constrained by our computational budget, all experiments use backbones of roughly 7B parameters, so the effect of structural supervision is characterized at a single scale rather than as a trend, and whether the reported gains persist or narrow with larger backbones remains open. We leave both questions, together with the cross-framework benchmark the first presupposes, to future work.

\newpage
\bibliographystyle{model1-num-names}
\bibliography{cas-refs}

\newpage

\bio{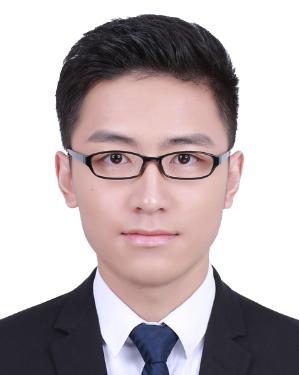}
\textbf{Minggui He} received the B.E. degree from Beijing University of Chemical Technology, China and M.E. from  Waseda University, Japan in 2016 and 2018 respectively. He is currently a Phd candidate in Waseda University. His research interests include Multi-modal content process and generation, large language models and image signal processing.
\endbio

\bio{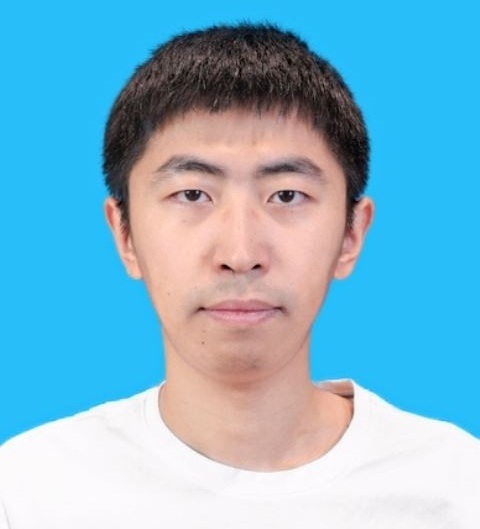}
\textbf{Mingchen Dai} is currently a third-year master’s student in Software Engineering at the University of Science and Technology of China. He received his B.E. degree from Qingdao University of Technology in 2021. His research interests include multi-modal large language models and chart-related analysis and generation.
\endbio

\bio{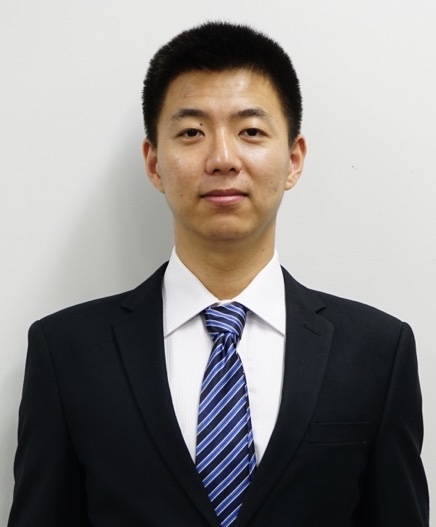}
\textbf{Jian Zhang} received the B.E. and M.E. degrees from The University of Electronic Science and Technology of China in 2016 and 2019 respectively. He was a Exchange Student with Waseda University, Japan, in 2017, and also got M.E. degree from Waseda University, Japan, in 2019. He has been a Ph.D. Student with Waseda University, Japan, since 2019. His research interests include computer vision, image processing and deep learning.
\endbio

\bio{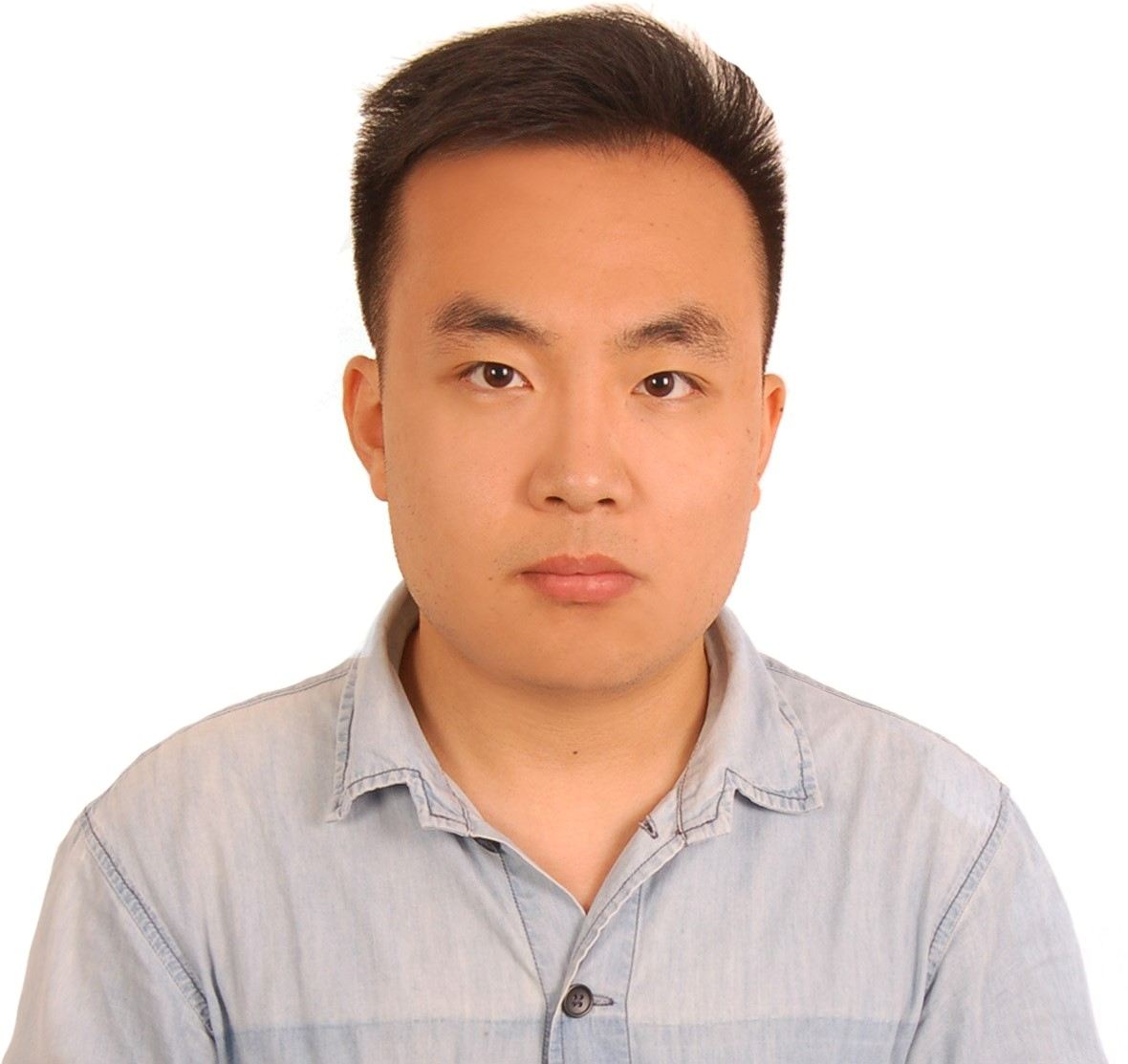}
\textbf{Yilun Liu} received the B.S. degree from Nankai University in 2020. He received the M.S. degree in Electrical and Computer Engineering from Georgia Institute of Technology in 2022. He is currently a Research Engineer with Huawei Translation Services Center in 2012 Lab, Huawei. His research interests include log analysis, large language models, machine translation, and natural language processing.
\endbio

\bio{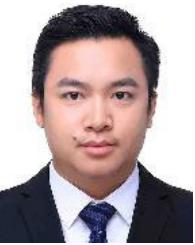}
\textbf{Shimin Tao} received the M.S. degree from Beijing University of Aeronautics and Astronautics, Beijing, China. He is currently a technology expert in Huawei 2012 laboratory. The main research direction is AIOps, machine translation, natural language processing, etc.
\endbio

\bio{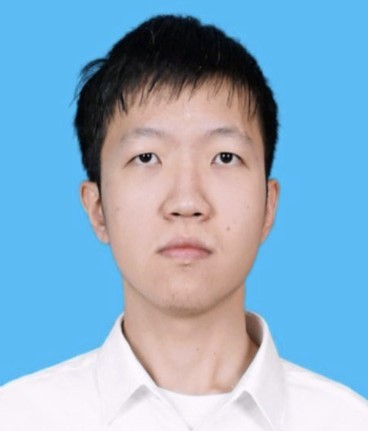}
\textbf{Pufan Zeng} received the B.E. degree from Beijing University of Chemical Technology, China. He is currently pursuing the master's degree with the University of Science and Technology of China. His research interests include large language models and natural language processing.
\endbio

\bio{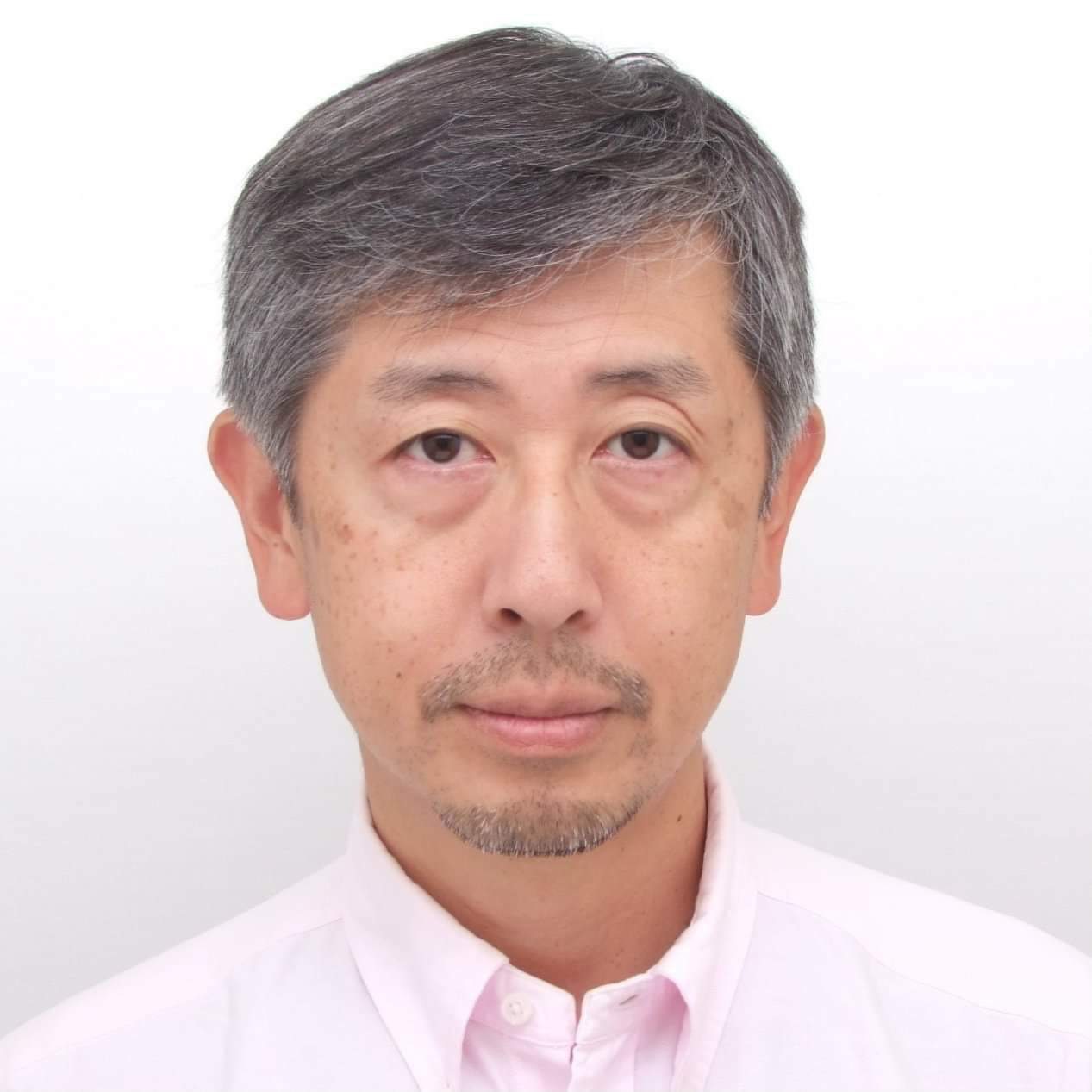}
\textbf{Osamu Yoshie} is currently a professor at Waseda University, Japan. His research interests include the analysis and synthesis of information so that a collective intelligence is generated in a community, which is composed of human, software and hardware. As an application of the community-based information processing, he is developing methodologies for consensus building in an organization, inspired from the research experience at FX Palo Alto Laboratory in USA and GMD (currently Fraunhofer) in Germany. He is a recipient of many awards from professional research societies, such as the Society of Instrument and Control Engineers (SICE), the Institute of Electrical Engineers (IEE) Japan, the Society of Plant Engineers Japan, Japan Industrial Management Association, etc. He is a fellow of IEE Japan.
\endbio

\bio{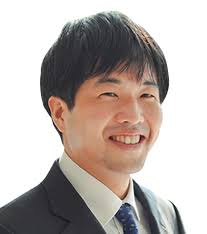}
\textbf{Yuya Ieiri} received the B.Eng., M.Eng., and Ph.D. degrees in Engineering from Waseda University, Japan, in 2017, 2019, and 2021, respectively. He is currently an Assistant Professor  at  Waseda University. His research interests include artificial intelligence, human–computer interaction, social systems engineering, augmented reality, tourism informatics, and social informatics.
\endbio

\end{document}